\documentclass[runningheads]{llncs}

\usepackage[T1]{fontenc}
\usepackage{microtype}
\usepackage{graphicx}
\usepackage{subfigure}
\usepackage{capt-of}
\usepackage{color}
\usepackage{cite}
\usepackage{tikz}
\usepackage[accsupp]{axessibility}  
\usepackage{amsmath,amssymb,amsfonts,dsfont,pifont,bm,bbm,mathrsfs,mathtools}
\usepackage{algorithm,algpseudocode,listings}
\usepackage{booktabs,multirow,adjustbox,diagbox}
\usepackage{xspace}
\usepackage[pagebackref=false,breaklinks=true,colorlinks=true,citecolor=blue,bookmarks=false]{hyperref}
\usepackage{cleveref}  
\crefname{section}{Sec.}{Secs.}
\Crefname{section}{Section}{Sections}
\crefname{table}{Tab.}{Tabs.}
\Crefname{table}{Table}{Tables}
\crefname{figure}{Fig.}{Figs.}
\Crefname{figure}{Figure}{Figures}
\crefname{equation}{Eq.}{Eqs.}
\Crefname{equation}{Equation}{Equations}
\newcommand{\LPIPS}{\textbf{LPIPS$\downarrow$}}  
\newcommand{\MSE}{\textbf{MSE$\downarrow$}}      
\newcommand{\Time}{\textbf{Time$\downarrow$}}    

\newcommand{\p}{{\rm\bf p}}         
\renewcommand{\P}{\mathcal{P}}      
\newcommand{\x}{{\rm\bf x}}         
\newcommand{\X}{\mathcal{X}}        
\newcommand{\z}{{\rm\bf z}}         
\newcommand{\Z}{\mathcal{Z}}        
\newcommand{\w}{{\rm\bf w}}         
\newcommand{\W}{\mathcal{W}}        
\newcommand{\WP}{\mathcal{W}^+}     
\newcommand{\E}{\mathbb{E}}         
\newcommand{\loss}{\mathcal{L}}     

\newcommand{\method}{PadInv\xspace}
\newcommand{\supp}{\textit{Supplementary Material}\xspace}
\makeatletter
\newcommand{\thankssymbol}[1]{\textsuperscript{\@fnsymbol{#1}}}
\makeatother

\newcommand{\blocks}[3]{\multirow{3}{*}{\(\left[\begin{array}{c}\text{1$\times$1, #2}\\[-.1em] \text{3$\times$3, #2}\\[-.1em] \text{3$\times$3, #1}\end{array}\right]\)$\times$#3}
}

\begin{document}

\pagestyle{headings}
\mainmatter
\def\ECCVSubNumber{877}

\title{High-fidelity GAN Inversion with Padding Space}

\titlerunning{High-fidelity GAN Inversion with Padding Space}
\author{
    Qingyan Bai\inst{1}\thanks{Equal contribution. \quad $^{\dagger}$ Corresponding author.},~
    Yinghao Xu\inst{2}\thankssymbol{1},~
    Jiapeng Zhu\inst{3},~
    Weihao Xia\inst{4},\\
    Yujiu Yang\inst{1\dagger},~ and 
    Yujun Shen\inst{5}
}

\authorrunning{Q. Bai et al.}

\institute{
    \textsuperscript{1}Shenzhen International Graduate School, Tsinghua University\quad
    \textsuperscript{2}CUHK\\
    \textsuperscript{3}HKUST\quad
    \textsuperscript{4}University College London\quad
    \textsuperscript{5}ByteDance Inc.\\
    \email{bqy20@mails.tsinghua.edu.cn}\quad
    \email{xy119@ie.cuhk.edu.hk}\\
    \email{\{jengzhu0, xiawh3\}@gmail.com}\\
    \email{yang.yujiu@sz.tsinghua.edu.cn}\quad
    \email{shenyujun0302@gmail.com}
}

\maketitle

\begin{abstract}

Inverting a Generative Adversarial Network (GAN) facilitates a wide range of image editing tasks using pre-trained generators.
Existing methods typically employ the latent space of GANs as the inversion space yet observe the insufficient recovery of spatial details.
In this work, we propose to involve the \textit{padding space} of the generator to complement the latent space with spatial information.
Concretely, we replace the constant padding (\textit{e.g.}, usually zeros) used in convolution layers with some instance-aware coefficients.
In this way, the inductive bias assumed in the pre-trained model can be appropriately adapted to fit each individual image.
Through learning a carefully designed encoder, we manage to improve the inversion quality both qualitatively and quantitatively, outperforming existing alternatives.
We then demonstrate that such a space extension barely affects the native GAN manifold, hence we can still reuse the prior knowledge learned by GANs for various downstream applications.
Beyond the editing tasks explored in prior arts, our approach allows a more flexible image manipulation, such as the separate control of face contour and facial details, and enables a novel editing manner where users can \textit{customize} their own manipulations highly efficiently.%
%
%
\footnote{Project page can be found \href{https://ezioby.github.io/padinv/}{here}.}

\end{abstract}
\section{Introduction}\label{sec:intro}

\begin{figure}[t]
    \centering
    \includegraphics[width=1.0\linewidth]{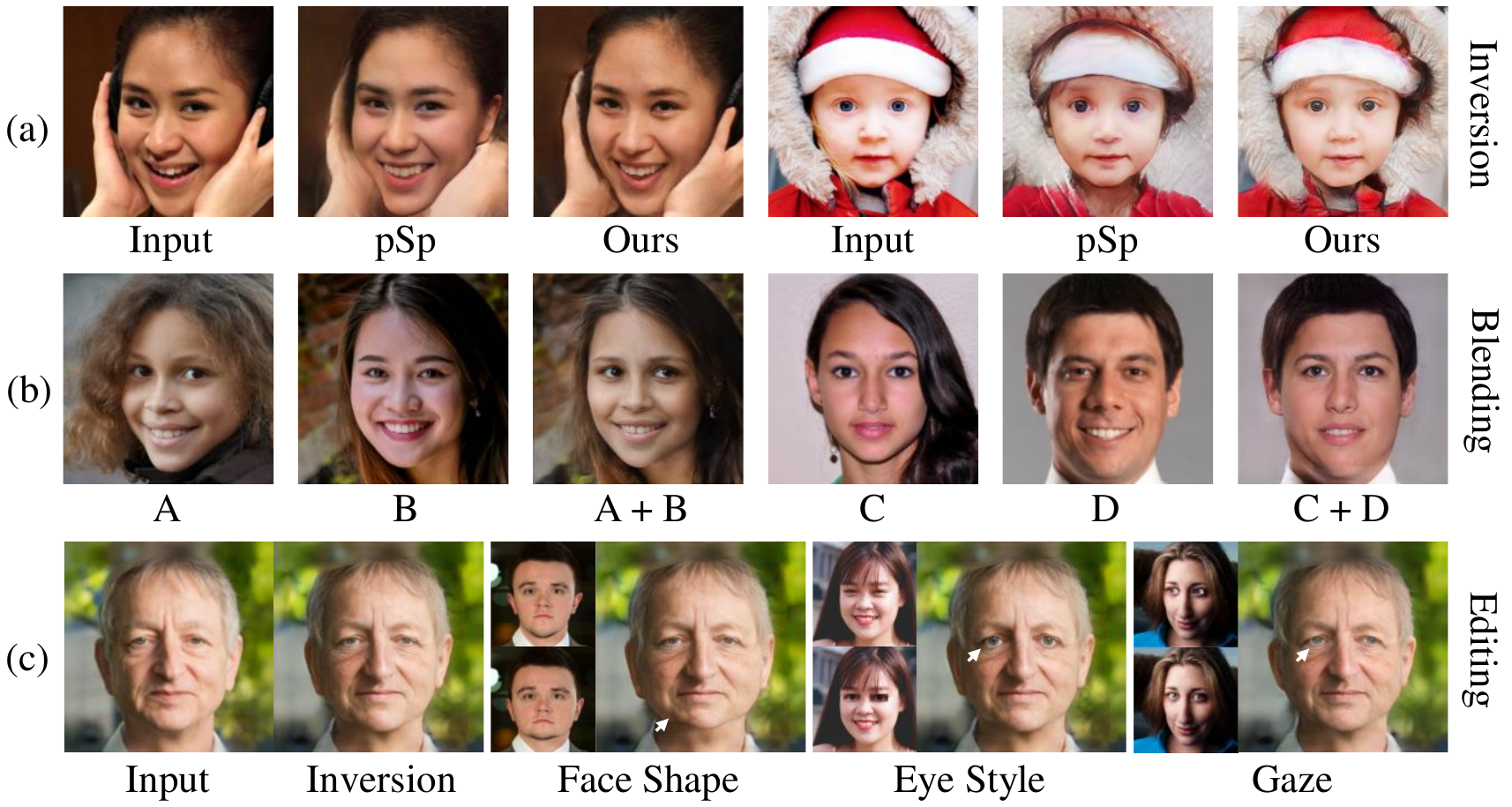}
    \caption{
        \textbf{Inversion and editing results} obtained by \method.
        (a) Our method better reconstructs the out-of-distribution objects (\textit{e.g.}, hands and hat) than pSp~\cite{richardson2021pSp}.
        (b) Introducing the padding space to complement the latent space allows separate control of face contour and facial details, facilitating face blending.
        (c) Versatile manipulations can be \textit{customized} with \textit{only one} image pair (\textit{i.e.}, on the left of each editing result)
    }
    \label{fig:teaser}
\end{figure}

Generative Adversarial Network (GAN)~\cite{goodfellow2014gan} has received wide attention due to its capability of synthesizing photo-realistic images~\cite{biggan, stylegan2, stylegan3}.
Recent studies have shown that GANs spontaneously learn rich knowledge in the training process, which can be faithfully used to control the generation~\cite{bau2018gandissection, shen2020interfacegan, jahanian2019steerability}.
However, the emerging controllability is hard to apply to real-world scenarios.
That is because the generator in a GAN typically learns to render an image from a randomly sampled latent code, and hence lacks the ability to make inferences on a target sample, limiting its practical usage.

The advent of GAN inversion techniques (\textit{i.e.}, inverting the generation process of GANs)~\cite{zhu2016generative, image2stylegan, zhu2020idinvert} appears to fill in this gap.
The core thought is to convert a given image to some GAN-interpretable representations~\cite{xu2021ghfeat}, which can be decoded by the generator to reconstruct the source.
That way, real image editing can be simply achieved by manipulating the inverted representations, reusing the pre-trained generator as a learned renderer.

The crux of GAN inversion is to find the appropriate representations that can recover the input as much as possible.
A common practice is to regularize the representations within the latent space of GANs~\cite{image2stylegan, zhu2020idinvert, richardson2021pSp, tov2021e4e}, which best matches the generation mechanism (\textit{i.e.}, the latent code uniquely determines the synthesis with the generator fixed).
However, merely using the latent space suggests unsatisfactory reconstruction performance.
The major reason causing such an issue is the inadequate recovery of some out-of-distribution objects, like the hands and hat in face images shown in \cref{fig:teaser}a.

In this work, we re-examine the procedure of how an image is produced, oriented to GANs with convolution-based generators, and get a deeper understanding of why some spatial details cannot be well recovered.
Recall that, to maintain the spatial dimensions of the input feature map, a convolution layer is asked to pad the feature map (\textit{e.g.}, usually with zeros) before convolution.
The padding is \textit{not learned} in the training phase and hence introduces some inductive bias~\cite{xu2021positional}, which may cause the ``texture-sticking'' problem~\cite{stylegan3}.
For instance, some padded constants may encode hair information such that hair gets stuck to some synthesized pixels however the latent code varies~\cite{stylegan3}.
Hereafter, when those pixels are filled with other objects (\textit{e.g.}, hat) in the target image, it becomes hard to invert them through parameter searching within the latent space.

To alleviate such a problem, we propose a high-fidelity GAN inversion approach, termed as \textbf{\method}, by \textit{incorporating the padding space of the generator as an extended inversion space} in addition to the latent space.
Concretely, we adapt the padding coefficients used in the generator for every individual image instead of inheriting the inductive bias (\textit{e.g.}, zero padding) assumed in GAN training.
Such an instance-aware reprogramming is able to complement the latent space with adequate spatial information, especially for those out-of-distribution cases.
It is noteworthy that the convolutional weights are still shared across samples, leaving the GAN manifold preserved.
Consequently, the prior knowledge learned in the pre-trained model is still applicable to the inversion results.
We also carefully tailor an encoder that is compatible with the newly introduced padding space, such that an image can be inverted accurately and efficiently.

We evaluate our algorithm from the perspectives of both inversion quality and image editing.
On the one hand, \method is capable of recovering the target image with far better spatial details than state-of-the-art methods, as exhibited in \cref{fig:teaser}a.
On the other hand, our approach enables two novel editing applications that have not been explored by previous GAN inversion methods.
In particular, we manage to control the image generation more precisely, including the \textit{separate manipulation of spatial contents and image style}.
As shown in \cref{fig:teaser}b, we achieve face blending by borrowing the contour from one image and facial details from another.
Furthermore, we come up with an innovative editing manner, which \textit{allows users to define their own editing attributes}.
As shown in \cref{fig:teaser}c, versatile manipulations are customized with \textit{only one} image pair, which can be created super efficiently either by graphics editors like Photoshop (\textit{e.g.}, face shape) or by the convenient copy-and-paste (\textit{e.g.}, eye style).

\section{Related Work}\label{sec:related}

\noindent\textbf{Generative Adversarial Networks (GANs).}
Recent years have witnessed the tremendous success of GANs in producing high-resolution, high-quality, and highly-diverse images~\cite{stylegan, biggan, stylegan2, stylegan3}.
Existing studies on GAN interpretation have affirmed that, pre-trained GAN models own great potential in a range of downstream applications, such as object classification~\cite{bigbigan, xu2021ghfeat}, semantic segmentation~\cite{zhang2021datasetgan}, video generation~\cite{stylegan3}, and image editing~\cite{shen2020interfacegan, shen2020sefa, gu2020image, yang2019HiGAN, jahanian2019steerability, ganspace, pan2020dgp, wu2020stylespace, zhu2021lowrankgan, ling2021editgan, bau2021paint, zhu2022resefa}.

\noindent\textbf{GAN Inversion.}
GAN inversion~\cite{zhu2016generative, xia2021ganinversion} aims at finding the reverse mapping of the generator in GANs.
Active attempts broadly fall into two categories, which are optimization-based~\cite{lipton2017precise, creswell2018inverting, image2stylegan, gu2020image, pan2020exploiting, huh2020ganprojection} and learning-based~\cite{perarnau2016invertible, lia, pidhorskyi2020alae, richardson2021pSp, xu2021ghfeat, alaluf2021restyle, tov2021e4e, zhu2016generative, zhu2020idinvert}.
However, they typically employ the native latent space of the generator as the inversion space yet observe inadequately recovered spatial details.
Some also~\cite{roich2021pivotal, alaluf2022hyperstyle, dinh2022hyperinverter} try to fine-tune the pre-trained generator, leading to better inversion but much slower speed.
To make GAN inversion spatially aware, some attempts~\cite{abdal2020image2stylegan++, stylegan2} incorporate the noise space of StyleGAN~\cite{stylegan}, while some concurrent studies~\cite{zhu2021barbershop, kang2021OOR, wang2021TengFei} propose to invert an image to an intermediate feature map such that only half of the generator will be used as the decoder.
By contrast, we involve the \textit{padding space} of the generator to complement the latent space with spatial information.
In doing so, the prior knowledge contained in the pre-trained model is barely affected, and hence our inversion results support various editing tasks well.

\noindent\textbf{Padding in Convolutional Neural Networks (CNNs).}
The side effect of constant padding used in CNNs has recently been studied~\cite{islam2019much, islam2021position, kayhan2020translation, alsallakh2020mind, xu2021ghfeat, stylegan3}.
It is revealed that padding would implicitly offer the absolute spatial location as the inductive bias~\cite{islam2019much, islam2021position, kayhan2020translation}, which may bring blind spots for object detectors~\cite{alsallakh2020mind}.
A similar phenomenon is also observed in generative models~\cite{xu2021positional}.
The unwanted spatial information would be utilized by the generator to learn fixed texture in some coordinates, also known as the ``texture-sticking'' problem~\cite{stylegan3}.
In this work, we manage to \textit{convert such a side effect to an advantage for GAN inversion} through replacing the constant padding assumed in the pre-trained generator with instance-aware coefficients.

\section{Method}\label{sec:method}

\begin{figure}[t]
    \centering
    \includegraphics[width=0.95\linewidth]{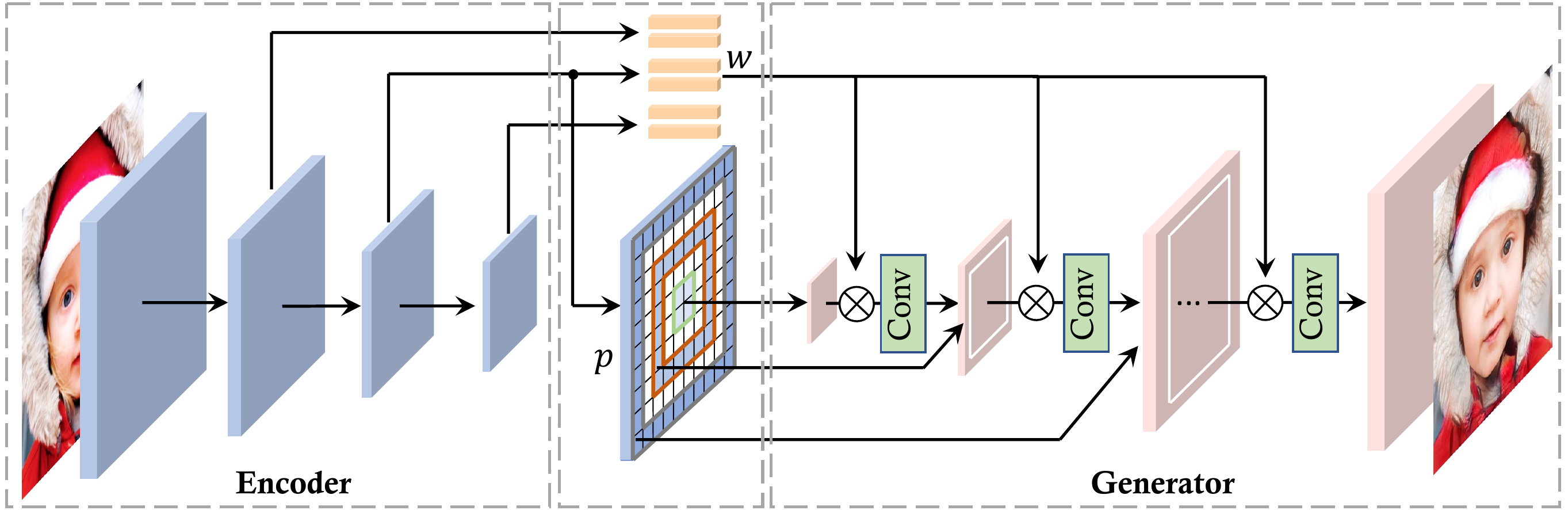}
    \caption{
        \textbf{Framework} of the proposed \method, which learns an encoder to invert the generation of a pre-trained convolution-based generator, $G(\cdot)$.
        Given a target image, our encoder not only maps it to layer-wise latent codes (\textit{i.e.}, $\w$), but also produces a coefficient tensor (\textit{i.e.}, $\p$) to \textit{replace the padding} used in the convolution layers of $G(\cdot)$.
        $\otimes$ denotes the AdaIN~\cite{adain} operation.
        The convolutional kernels of $G(\cdot)$ are \textit{fixed} in the training phase and \textit{shared} across all samples
    }
    \label{fig:framework}
\end{figure}

\subsection{Preliminaries}

The StyleGAN family~\cite{stylegan, stylegan2, stylegan3} has significantly advanced image generation of convolution-based GANs.
With the start-of-the-art synthesis performance, they are widely used for GAN inversion studies~\cite{image2stylegan, richardson2021pSp, xu2021ghfeat, alaluf2021restyle, tov2021e4e, zhu2020idinvert, abdal2020image2stylegan++, zhu2021barbershop, kang2021OOR, wang2021TengFei}.
Like most GAN variants~\cite{goodfellow2014gan}, the generator of StyleGAN takes a latent code, $\z$, as the input and outputs an image, $\x=G(\z)$.
Differently, StyleGAN~\cite{stylegan} learns a more disentangled latent space, $\W$, on top of the original latent space, $\Z$, and feeds the disentangled latent code, $\w$, to every single convolution layer through AdaIN~\cite{adain}.
Prior arts have verified that $\WP$ space~\cite{image2stylegan}, which is an extension of $\W$ via repeating $\w$ across layers, yields the most promising results for GAN inversion.
However, existing approaches still suffer from the insufficient recovery of some spatial details, like hats and earrings within face images.
We attribute such an issue to the fact that AdaIN acts on the feature map globally~\cite{adain, stylegan}, therefore, the latent code fails to provide adequate spatial information for image reconstruction.
In this work, we propose an innovative solution, which complements the latent space with the padding space of the generator.

\subsection{Padding Space for GAN Inversion}\label{subsec:padding-space}

Recall that the StyleGAN generator~\cite{stylegan, stylegan2} stacks a series of convolution layers (\textit{i.e.}, with $3\times3$ kernel size) for image generation.
For each layer, it pads the input feature map before convolution to keep its spatial dimensions unchanged.
Zero padding is commonly adopted and frozen in GAN training.
Recent studies~\cite{xu2021ghfeat, stylegan3} have pointed out that, the constant padding acts as an inductive bias through exposing the spatial location to the generator~\cite{xu2021ghfeat}, and hence causes the ``texture-sticking'' problem~\cite{stylegan3}.
In other words, some coordinates may \textit{get stuck with fixed texture} no matter what latent code is given~\cite{stylegan3}.
Such a side effect of padding complicates the inversion task, especially when the target image contains out-of-distribution objects (\textit{e.g.}, a hat appears at the pixels that are supposed to be hair as encoded by padding).
From this perspective, merely employing the latent space as the inversion space is highly insufficient for a fair reconstruction.

To tackle the obstacle, we propose to \textit{adapt the padding coefficients} used in the generator for high-fidelity GAN inversion.
That way, the inductive bias offered in the generator can be appropriately adjusted to fit every individual target.
We call the parameter space, which is constructed by those instance-aware convolutional paddings, as the padding space, $\P$.
Intuitively, $\P$ is incorporated as an extended inversion space to complement the latent space (\textit{i.e.}, $\WP$) with spatial information.
Besides, the StyleGAN generator learns image synthesis starting from a constant tensor~\cite{stylegan}.
The constant input can be viewed as a special padding (\textit{i.e.}, from a tensor with shape $0\times0$) without performing convolution.
In our approach, the initial constant is also adapted for each target sample.

\subsection{Encoder Architecture}\label{subsec:encoder}

To accelerate the inversion speed, we learn an encoder, $E(\cdot)$, following prior works~\cite{zhu2020idinvert, richardson2021pSp, alaluf2021restyle, tov2021e4e}.
The architecture of the encoder is carefully tailored to get compatible with the newly introduced padding space, $\P$.
In particular, in addition to the layer-wise latent codes, $\{\w_\ell\}_{\ell=1}^{L}$, our encoder also predicts a coefficient map from the input, as shown in \cref{fig:framework}.
Here, $L$ stands for the total number of layers in the generator.
Values from the coefficient map will be used to replace the paddings used in $G(\cdot)$, denoted as $\{\p_\ell\}_{\ell=0}^{L}$, at the proper resolution.
$\p_0$ represents the initial constant input.
For example, the coefficients at the outer of the central $18\times18$ patch are borrowed as the padding for the convolution at $16\times16$ resolution.
Note that, we only apply one set of coefficients for each resolution to avoid the two convolution layers of the same resolution from sharing the same padding.
In practice, we replace $\p_0, \p_1, \p_3, \p_5, \dots$, keeping the padding for other layers untouched.

Our encoder backbone is equipped with several residual blocks~\cite{resnet}, and a Feature Pyramid Network (FPN)~\cite{lin2017fpn} is adopted to inspect the input at multiple levels.
The outputs of the last three blocks (\textit{i.e.}, with the smallest resolutions) are employed to learn the latent codes, $\{\w_\ell\}_{\ell=1}^{L}$, in a hierarchical manner.
They correspond to the generator layers in reverse order.
Taking an 18-layer StyleGAN generator as an example, the last block produces latent codes for layers 1-3, the second last for layers 4-7, while the third last for layers 8-18, respectively.
Meanwhile, we also choose a max resolution\footnote{We empirically verify that 32 is the best choice in \cref{subsec:inversion}.} where padding replacement will be performed.
The coefficient map is projected (\textit{i.e.}, implemented with a convolution layer with $1\times1$ kernel size) from the FPN feature with target resolution, and then properly resized to allow padding in the generator.
For instance, given the max resolution as 32, we convolve the $32\times32$ FPN feature and upsamples the convolution result to $34\times34$, which is the minimum size to pad $32\times32$ features.
More details can be found in \supp.

\subsection{Training Objectives}\label{subsec:loss}

\noindent\textbf{Encoder}.
We develop the following loss functions for encoder training.
\begin{description}
    \item\textit{1. Reconstruction loss.}
    Image reconstruction is the primary goal of GAN inversion.
    Hence, we optimize the encoder with both pixel guidance and perceptual guidance, as
    \begin{align}
        \loss_{pix} &= || \x - G(E(\x)) ||_2,  \label{eq:pixel_loss} \\ 
        \loss_{per} &= || \phi(\x) - \phi(G(E(\x))) ||_2,  \label{eq:perceptual_loss}
    \end{align}
    where $||\cdot||_2$ denotes the $l_2$ norm.
    $\phi(\cdot)$ is a pre-trained perceptual feature extractor~\cite{zhang2018lpips}, which is popularly used for GAN inversion~\cite{zhu2020idinvert, richardson2021pSp}.

    \item\textit{2. Identity loss.}
    This loss function is particularly designed for inverting face generation models, following \cite{richardson2021pSp}.
    It can help the encoder to focus more on the face identity, which is formulated as
    \begin{align}
        \loss_{id} = 1 - \cos(\psi(\x), \psi(G(E(\x)))),  \label{eq:id_loss}
    \end{align}
    where $\cos(\cdot, \cdot)$ computes the cosine similarity.
    $\phi(\cdot)$ is a pre-trained identity feature extractor~\cite{deng2019arcface}.
    This loss will be \textit{disabled} for non-face models.

    \item\textit{3. Adversarial loss.}
    To avoid blurry reconstruction~\cite{richardson2021pSp}, the encoder is also asked to compete with the discriminator, resulting in an adversarial training manner. The adversarial loss is defined as
    \begin{align}
        \loss_{adv} = - \underset{\x \sim \X} \E[ D(G(E(\x))) ],  \label{eq:adv_loss}
    \end{align}
    where $\E[\cdot]$ and $\X$ are the expectation operation and the real data distribution, respectively.
    $D(\cdot)$ stands for the discriminator.

    \item\textit{4. Regularization loss.}
    Recall that StyleGAN generator maintains an averaged latent code, $\overline{\w}$, which can be viewed as the statistics of the latent space $\W$.
    We expect the inverted code to be subject to the native latent distribution as much as possible~\cite{tov2021e4e}.
    Hence, we regularize it with
    \begin{align}
        \loss_{reg} = || E_{latent}(\x) - \overline{\w} ||_2,  \label{eq:reg_loss}
    \end{align}
    where $E_{latent}(\cdot)$ indicates the latent part of the encoder, excluding padding.
\end{description}
To summarize, the full objective for our encoder is:
\begin{align}
    \loss_{E} = \lambda_{pix} \loss_{pix} +
                \lambda_{per} \loss_{per} +
                \lambda_{id} \loss_{id} +
                \lambda_{adv} \loss_{adv} +
                \lambda_{reg} \loss_{reg},
    \label{eq:encoder-loss}
\end{align}
where $\lambda_{pix}$, $\lambda_{per}$, $\lambda_{id}$, $\lambda_{adv}$, and $\lambda_{reg}$ are loss weights to balance different terms.

\noindent\textbf{Discriminator}.
The discriminator is asked to compete with the encoder, as
\begin{align}
    \loss_{D} = \underset{\x \sim \X} \E[ D(G(E(\x))) ]
              - \underset{\x \sim \X} \E[ D(\x) ]
              - \frac{\gamma}{2} \underset{\x \sim \X} \E[ ||\nabla_{\x}D(\x)||_2 ], \label{eq:discriminator-loss}
\end{align}
where $\gamma$ is the hyper-parameter for gradient penalty~\cite{wgan-gp}.

\section{Experiments}\label{sec:exp}

In this part, we conduct extensive experiments to evaluate the effectiveness of the proposed \method.
\cref{subsec:exp-setting} introduces the experimental settings, such as  datasets and implementation details.
In~\cref{subsec:inversion}, we experimentally show the superiority of the proposed method~\method in terms of inversion quality and real image editing with off-the-shelf directions. We also conduct ablation studies.
In \cref{subsec:property}, we explore the property of the proposed padding space, which enables separate control of spatial contents and image style.
A face blending application is introduced to further present the superiority of the padding space. 
\cref{subsec:application2_1ShotEdit} shows a novel editing method that could be achieved by providing customized image pairs. We also quantitatively evaluate the editing effects of semantic directions in $\W+$ space and $\P$ space.

\subsection{Experimental Settings}\label{subsec:exp-setting}

We conduct experiments on the high-quality face datasets FFHQ~\cite{stylegan} and CelebA-HQ~\cite{liu2015deep,progan} for face inversion. 
We use the former 65k FFHQ faces as the training set (the rest 5k for further evaluation) and test set of CelebA-HQ for quantitative evaluation.
For scene synthesis, we adopt LSUN Church and Bedroom~\cite{yu2015lsun} and follow the official data splitting strategy.
The GANs to invert are pre-trained following StyleGAN~\cite{stylegan}.
As for the loss weights, we set $\lambda_{id}$ = 0.1 and $\lambda_{adv}$ = 0 for encoders trained on FFHQ, $\lambda_{adv}$ = 0.03 and $\lambda_{id}$ = 0 for encoders trained on LSUN.
For all the experiments, we set $\lambda_{pix}$ = 1, $\lambda_{per}$ = 0.8, $\lambda_{reg}$ = 0.003, and $\gamma$ = 10.
Our codebase is built on Hammer~\cite{hammer}.

\setlength{\tabcolsep}{5pt}
\begin{table*}[!t]
  \caption{
    \textbf{Quantitative comparisons} between different inversion methods on three datasets, including FFHQ (human face)~\cite{stylegan}, LSUN Church (outdoor scene)~\cite{yu2015lsun}, and LSUN Bedroom (indoor scene)~\cite{yu2015lsun}
  }
  \label{tab:reconstruction_quat}
  \scriptsize\centering
  \begin{tabular}{l|ccc|cc|cc}
                \toprule
                & \multicolumn{3}{c|}{\textbf{Face}}
                & \multicolumn{2}{c|}{\textbf{Church}} 
                & \multicolumn{2}{c}{\textbf{Bedroom}}                       \\
                \midrule
                & \MSE  & \LPIPS  & \Time
                &    \MSE    &    \LPIPS
                &    \MSE    &    \LPIPS          \\ 
                \midrule
    StyleGAN2~\cite{stylegan2}
                &  0.020  &   0.09  & 122.39
                &  0.220  &   0.39   
                &  0.170  &   0.42          \\
                \midrule
    ALAE~\cite{pidhorskyi2020alae}
                &  0.15  &  0.32  &  \textbf{0.0208}
                &   -    &    -         
                &  0.33  &  0.65        \\
    IDInvert~\cite{zhu2020idinvert}
                &   0.061   &    0.22   &  0.0397 
                &   0.140   &    0.36   
                &   0.113   &    0.41          \\
    pSp~\cite{richardson2021pSp}
                &  0.034  &  0.16  &  0.0610
                &   0.127    &    0.31  
                &   0.099    &    0.34        \\
    e4e~\cite{tov2021e4e}
                &  0.052  &  0.20   &  0.0610
                &  0.142   &  0.42   
                &   -        &    -         \\
    $\text{Restyle}_{pSp}$~\cite{alaluf2021restyle}
                &  0.030  &  0.13   &  0.2898
                &   0.090        &    0.25  
                &   -        &    -          \\
    $\text{Restyle}_{e4e}$~\cite{alaluf2021restyle}
                &  0.041  &  0.19   & 0.2898 
                &  0.129        &    0.38
                &   -        &    -        \\
    $\text{Ours}$
                & \textbf{0.021}  & \textbf{0.10}  & 0.0629
                &   \textbf{0.086}    &    \textbf{0.22} 
                &   \textbf{0.054}   &    
                \textbf{0.21}          \\ 
                \bottomrule
  \end{tabular}
\end{table*}

\begin{figure}[!ht]
    \centering
    \includegraphics[width=1.0\linewidth]{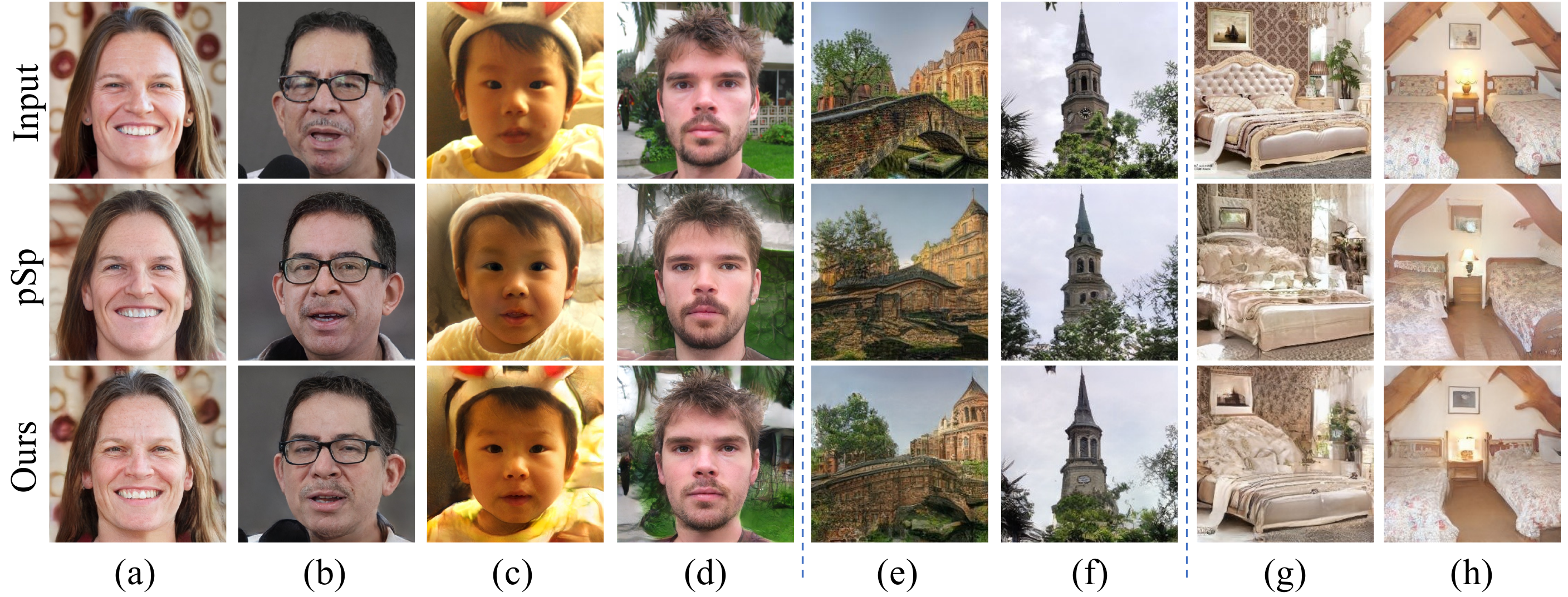}
    \caption{
        \textbf{Qualitative comparisons} between our approach and the baseline, pSp~\cite{richardson2021pSp}, where we better recover the spatial details of the input images
    }
    \label{fig:inversion}
\end{figure}

\subsection{GAN Inversion Performance}\label{subsec:inversion}

\noindent\textbf{Quantitative Results}.
We quantitatively evaluate the quality of the reconstructed images in \cref{tab:reconstruction_quat}.
Pixel-wise MSE and perceptual LPIPS~\cite{zhang2018lpips} are adopted as metrics to evaluate the reconstruction performance. 
We also include the inference time of face inversion to measure model efficiency.
As shown in \cref{tab:reconstruction_quat}, 
our approach leads to a significant improvement on face images compared with encoder-based baselines.
It even achieves a comparable performance to the optimization-based StyleGAN2 with better efficiency. 
When applied to more challenging datasets $\textit{e.g.}$, church and bedroom, StyleGAN2 fails to reconstruct images well because $\overline{\w}$ as the starting point for optimization cannot handle the unaligned data with large spatial variations.
However, our method is robust and still outperforms all baselines.
It demonstrates that the proposed padding space can provide fine-grained spatial details of images, resulting in better reconstruction quality.

\noindent\textbf{Qualitative Results}.
In \cref{fig:inversion}, we compare the reconstructed images of our method with the baseline method pSp~\cite{richardson2021pSp} on the face, church, and bedroom.
Even though pSp can restore the foreground of the given face images, it struggles to reconstruct spatial details, as in \cref{fig:inversion} (a), (b), (c), (d).
Thanks to the proposed padding space, our approach enhances the spatial modeling capacity of the inversion space so that the fine-grained background texture, microphone, and headwear can be reconstructed well, and the identity can be preserved.    
When it comes to challenging church and bedroom datasets with large spatial variations, pSp can achieve satisfying performance on restoring the texture but still struggles to reconstruct the spatial structure of the given images, as shown in \cref{fig:inversion} (e).
Our method can preserve structural details better because the proposed padding space can complement the original $\W+$ space with the instance-aware padding coefficients.
For instance, the structural information of buildings, such as the number of spires, is preserved more. The bridge, sheet and painting in \cref{fig:inversion}(e), (g), (h) are also well reconstructed.

\noindent\textbf{Ablation Studies}.
The proposed padding space is adopted to complement the $\W+$ space, and thus its scale is critical to the quality of the reconstructed image.
We adjust its scale by varying the padding layers.
\cref{tab:ablation} shows the reconstruction performance on the test sets of CelebA-HQ and LSUN Church.
We use pSp~\cite{richardson2021pSp} as the baseline, which only utilizes $\W+$ space for inversion. 
We firstly extend the inversion space with $\p_{0}$, which represents the constant input of the generator. 
The performance is slightly improved due to the introduction of the case-specific information.
Then we progressively enlarge the padding space by predicting padding coefficients for more layers as shown in \cref{fig:framework}.
As the padding space increases, the reconstruction performance becomes better.
This demonstrates that more padding coefficients can provide more precise information for spatial modeling, resulting in the improvement of image reconstruction. 
When padding layers come to 9 (resolution of 64$\times$64), the inversion performance gets a drop. 
We assume that the parameter of the padding space is too large, leading to overfitting on the training set. 
Thus the largest index of padding layers is chosen as 7 (resolution of 32$\times$32) to maintain the reconstruction performance.
We also study the effects of the regularization loss $\mathcal{L}_{reg}$. 
This regularization is designed to encourage the inverted code in $\W+$ space to be subject to the native latent distribution.
Equipping $\mathcal{L}_{reg}$ to the model leads to a slight drop in reconstruction performance, but the editing ability becomes better as in \supp.

\setlength{\tabcolsep}{12pt}
\begin{table*}[t]
  \caption{
      \textbf{Ablation studies} on the number of layers to replace padding, as well as the regularization loss, $\mathcal{L}_{reg}$, introduced in \cref{eq:reg_loss}
  }
  \label{tab:ablation}
  \scriptsize\centering
  \begin{tabular}{l|cc|cc}
                \toprule
                & \multicolumn{2}{c|}{\textbf{Face}} 
                & \multicolumn{2}{c}{\textbf{Church}}                       \\
                \midrule
                &    \MSE    &    \LPIPS  &  \MSE  & \LPIPS \\
                \midrule
      Baseline
                &   0.0344    &   0.161
                &   0.1272    &    0.311 \\
      $+ \p_{0}$
                &   0.0343    &   0.156
                &   0.1257    &    0.307 \\
      $+ \p_{0, 1}$
                &  0.0341    &    0.154
                &  0.1241    &    0.309 \\
      $+ \p_{0, 1, 3}$
                &  0.0252  & 0.115    
                &   0.1005  & 0.253 \\
      $+ \p_{0, 1, 3, 5}$
                &  0.0190  &  0.091
                &  0.0891  &  0.218 \\
      $+ \p_{0, 1, 3, 5, 7}$
                &   0.0186    &    0.090
                &   0.0838    &    0.209 \\
      $+ \p_{0, 1, 3, 5, 7, 9}$
                &   0.0216    &    0.108
                &   0.0979    &    0.271 \\
        \midrule
      $+ \p_{0, 1, 3, 5, 7}$ \& $\loss_{reg}$
                &   0.0214    &    0.103
                &   0.0866    &    0.222 \\
                \bottomrule
  \end{tabular}
\end{table*}

\begin{figure}[t]
    \centering
    \includegraphics[width=0.9\linewidth]{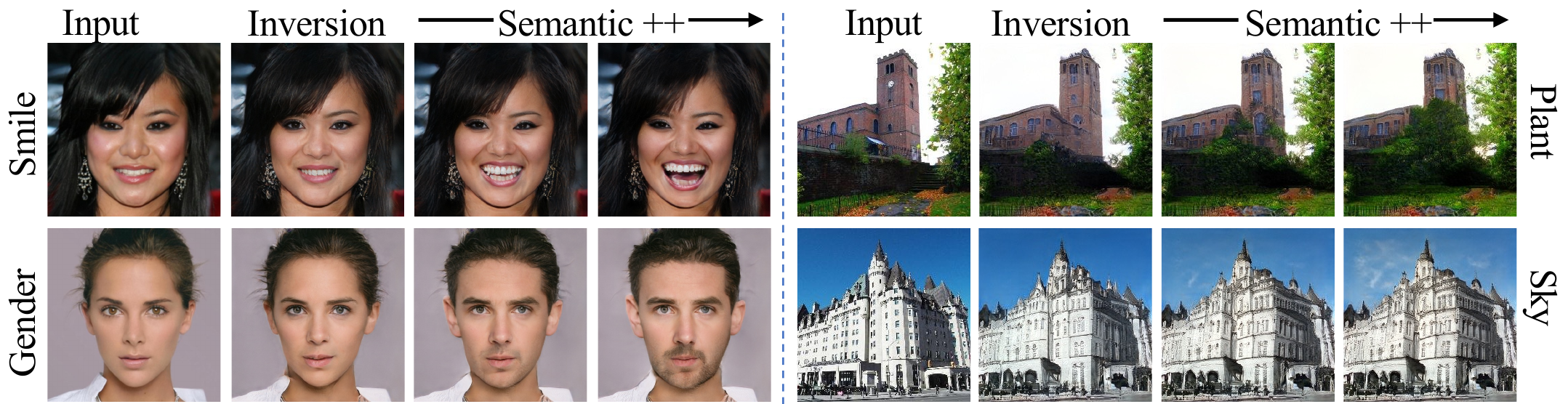}
    \caption{
        \textbf{Real image editing} with \textit{off-the-shelf} semantics identified by InterFaceGAN~\cite{shen2020interfacegan} (smile and gender), HiGAN~\cite{yang2019HiGAN} (plant), and LowRankGAN~\cite{zhu2021lowrankgan} (sky)
        }
    \label{fig:editing}
\end{figure}

\noindent\textbf{Editability}.
Having verified our method in better reconstruction quality, here we explore its editability.
We utilize off-the-shelf semantic directions from \cite{shen2020interfacegan, zhu2021lowrankgan, yang2019HiGAN} to edit the inversion results.
\cref{fig:editing} presents the results of manipulating faces and churches.
Our method can preserve most other details when manipulating a particular facial attribute. 
For churches, the semantics changes smoothly on the high-fidelity reconstructed images.
These editing results support that the extra padding space can not only invert the given images in high quality but also facilitate them with good properties on image manipulation.

\begin{figure}[!t]
    \centering
    \includegraphics[width=0.95\linewidth]{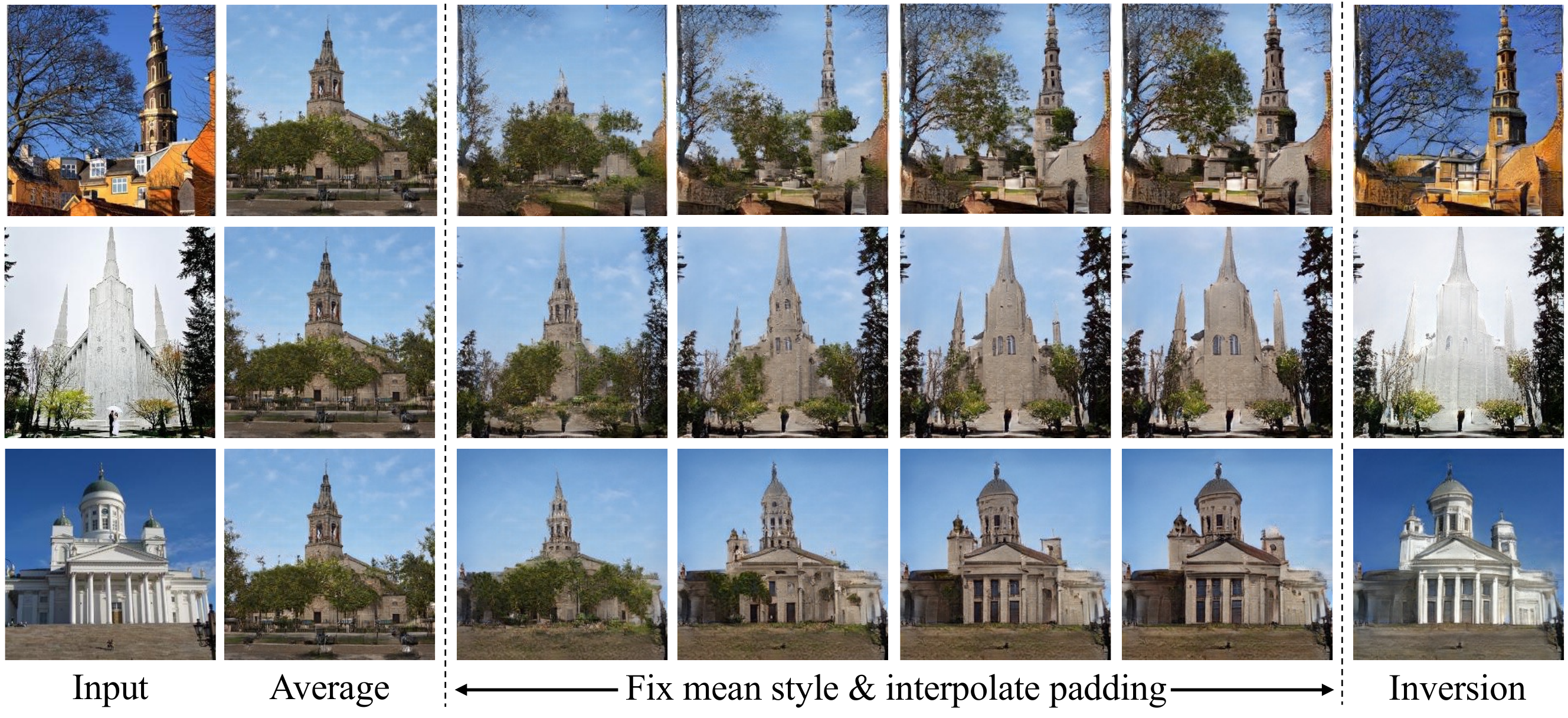}
    \caption{
        \textbf{Analysis of the padding effect} on LSUN Church~\cite{yu2015lsun}.
        We fix the latent code as the statistical average and interpolate the padding from the fixed constants in the generator to the coefficients specifically learned for inversion.
        It verifies that padding encodes the spatial information
    }
    \label{fig:church_inter1Pos}
\end{figure}

\begin{figure}[t]
    \centering
    \includegraphics[width=0.95\linewidth]{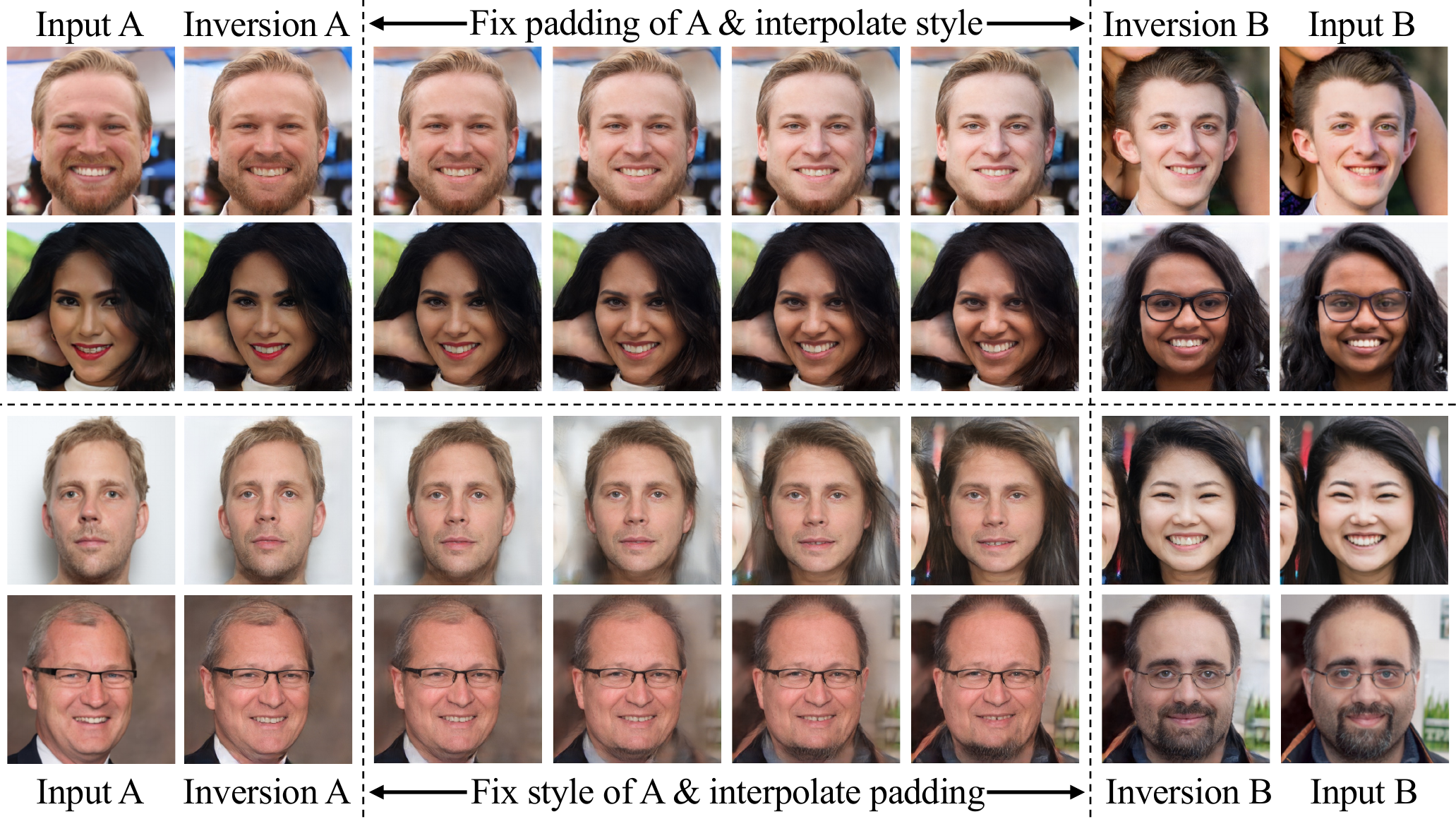}
    \caption{
        \textbf{Analysis of the extended inversion space} on FFHQ~\cite{stylegan}.
        We perform interpolation both in the latent space and in the padding space.
        It turns out that the latent and padding tends to encode style and spatial information, respectively
    }
    \label{fig:face_inter2}
\end{figure}

\begin{figure}[!ht]
    \centering
    \includegraphics[width=0.95\linewidth]{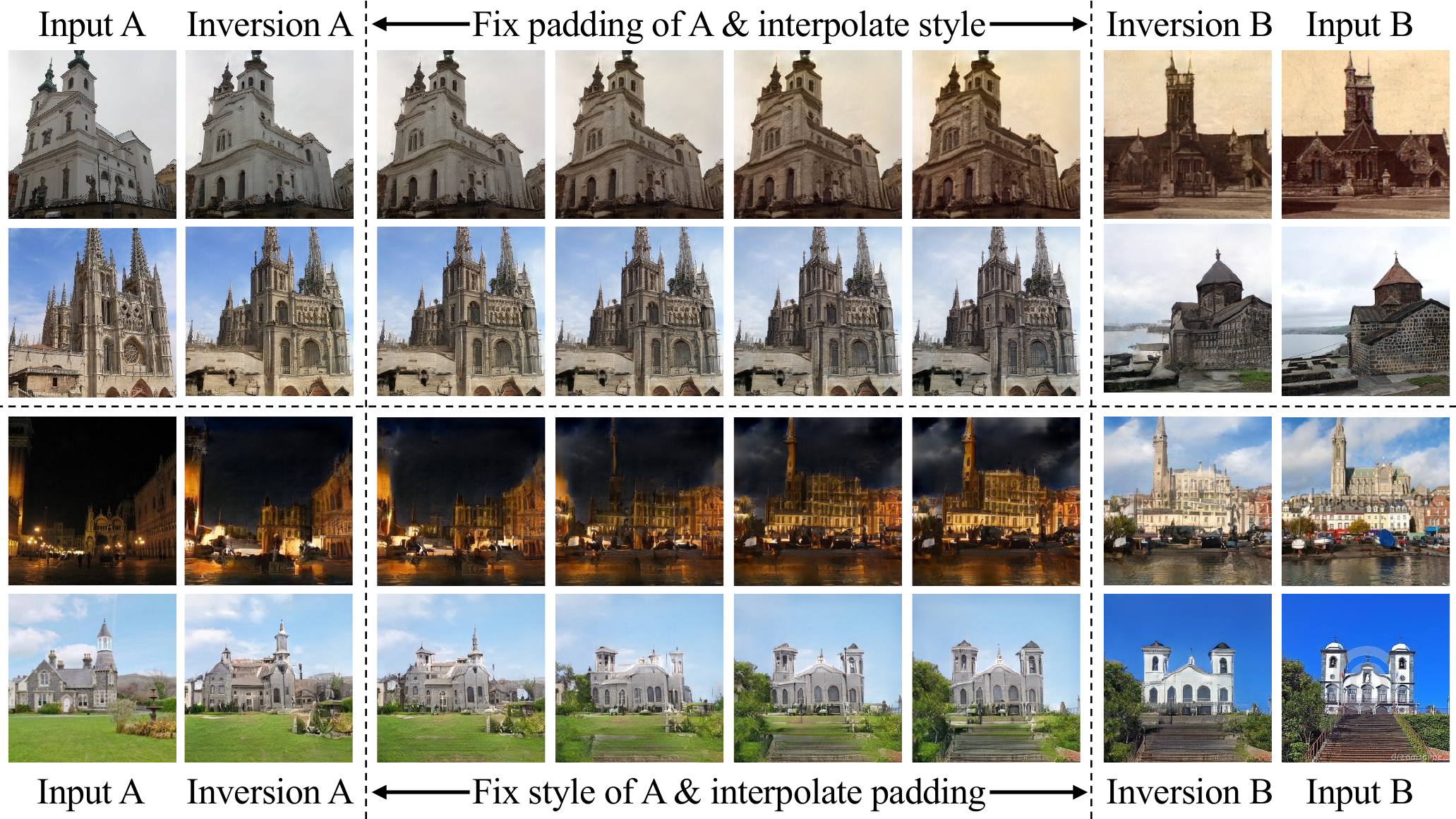}
    \caption{
        \textbf{Analysis of the extended inversion space} on LSUN Church~\cite{yu2015lsun}.
        We perform interpolation both in the latent space and in the padding space
    }
    \label{fig:church_inter2}
\end{figure}

\begin{figure}[!ht]
    \centering
    \includegraphics[width=0.95\linewidth]{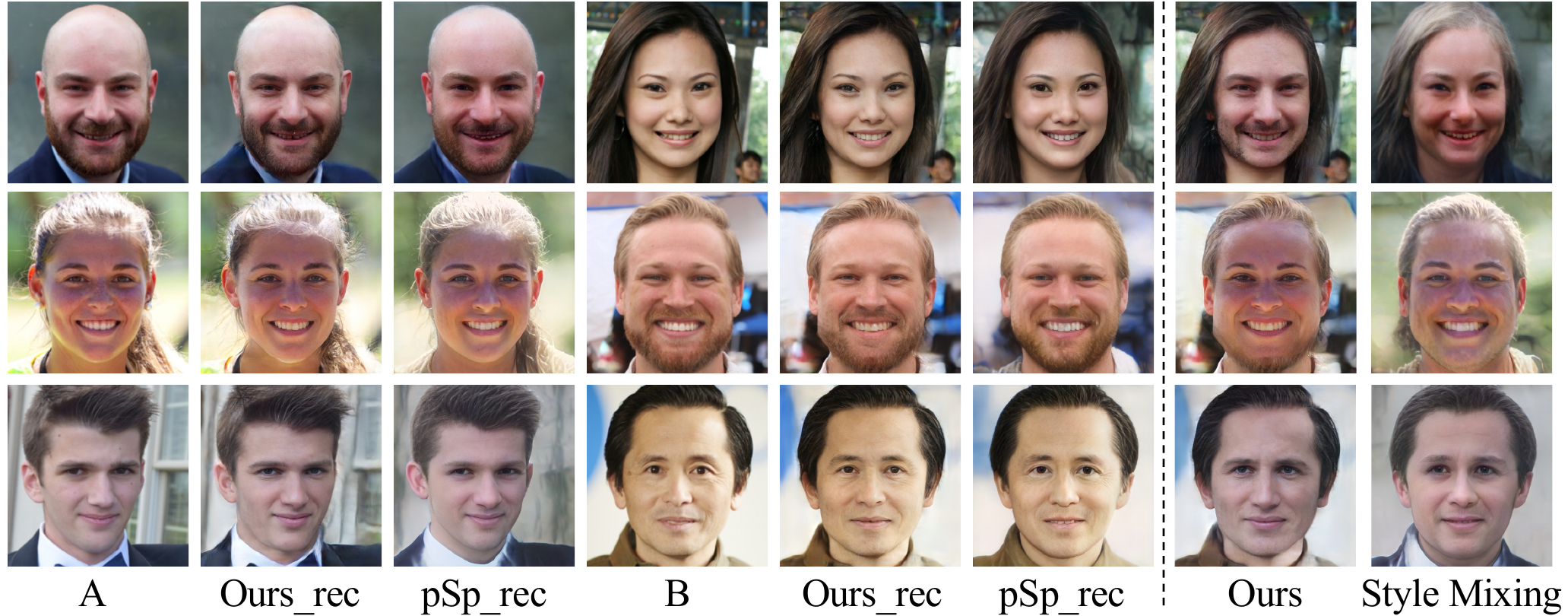}
    \caption{
       \textbf{Face blending results} by borrowing the inverted latent of A and the inverted padding of B.
       Different from style mixing, our approach enables the separate control of the facial details (\textit{e.g.}, the features of A), and the face contour (B)
    }
    \label{fig:face_blending}
\end{figure}

\subsection{Separate Control of Spatial Contents and Image Style}\label{subsec:property}

\noindent\textbf{Property of Padding Space.}
In this part, we investigate the proposed padding space's properties using interpolation experiments on the FFHQ and LSUN test sets.
We first conduct interpolation experiments between a real image and the average image synthesized with the average latent code $\overline{\w}$ and original paddings.
The real image is inverted with our encoder into the style latent codes and padding coefficients.
The paddings between the average image and the inversion result are then interpolated, while the latent codes from $\W+$ space are kept as $\overline{\w}$. 
\cref{fig:church_inter1Pos} demonstrates the interpolation results on the test sets of LSUN Church.
The results of interpolation on FFHQ and LSUN Bedroom are shown in \supp.
The style of the interpolated images is nearly identical, while the spatial structure is gradually transformed from the average image to the given image.
It suggests that the learned padding coefficients encode the structural information of the given images.  

Interpolation experiments between two real images are further conducted to validate the separate controllability of the spatial contents and image style.
We first invert two real images A and B to $\P$ and $\W+$ space.
Then we fix one of the paddings or latent codes and interpolate the other.
As illustrated in \cref{fig:face_inter2} and \cref{fig:church_inter2}, when padding coefficients are kept to be fixed, the style varies smoothly when the latent code changes.
Interpolating paddings with fixed style latent codes leads to the results with the same style and different spatial structures.
The interpolation results on LSUN Bedroom can be found in \supp.
Above interpolation results indicate that the paddings of $\P$ space encode structure information such as the pose and shape, which is complementary to style information encoded in latent codes of $\W+$ space.
We can leverage the properties of two spaces to achieve the independent control of spatial contents and image style.

\noindent\textbf{Application on Face Blending.}
As discussed in \cref{subsec:property}, paddings of $\P$ space control the spatial structure of the image. 
For face images, it controls the face shape and background.
Based on this property, we can naturally achieve face blending which integrates the shape and background of one face and facial features of another.
Specifically, we invert a real source face A and target portrait B by the encoder.
Then the blended face can be synthesized by sending the latent codes of A and padding coefficients of B to the pre-trained generator.
In \cref{fig:face_blending}, we compare our face blending results with naively mixing the latter 11 style latent codes as pSp~\cite{richardson2021pSp}.
Obviously, the naive style mixing in pSp introduces redundant information of the source image A and cannot keep the background and hair of the target image B.
However, our method can transfer the identity information of face A well and inherit the structure of target face B perfectly.
It is worth noting that, the whole face blending process does not require any pre-trained segmentation or landmark models. 
It leverages the structure information encoded in $\mathcal{P}$ space, enabling the independent control of the face structure and details. 

\subsection{Manipulation with Customized Image Pair}\label{subsec:application2_1ShotEdit}

\noindent\textbf{Implementation.}
Based on the padding space, we propose a novel editing method.
We find the semantic direction defined by one pair of customized image labels can be successfully applied to arbitrary samples. 
The paired images can be obtained by human retouching or naive copy-and-paste.
Specifically, we firstly invert the paired label images A and A' into $\P$ and $\W+$ space. 
The difference of the padding coefficients $\mathbf{n_{p}}$ in $\P$ space and difference of style codes $\mathbf{n_{s}}$ in $\W+$ space can be viewed as the semantic directions defined by the given paired images.
Then other real images can be edited using $\mathbf{n_{p}}$ and $\mathbf{n_{s}}$.
Considering that the paddings and latent codes encode various information, we propose to apply the semantic direction of one (either $\P$ or $\W+$) space, to maintain the editing performance.

\begin{figure}[t]
    \centering
    \begin{minipage}[b]{0.43\linewidth}
        \includegraphics[width=1.0\linewidth]{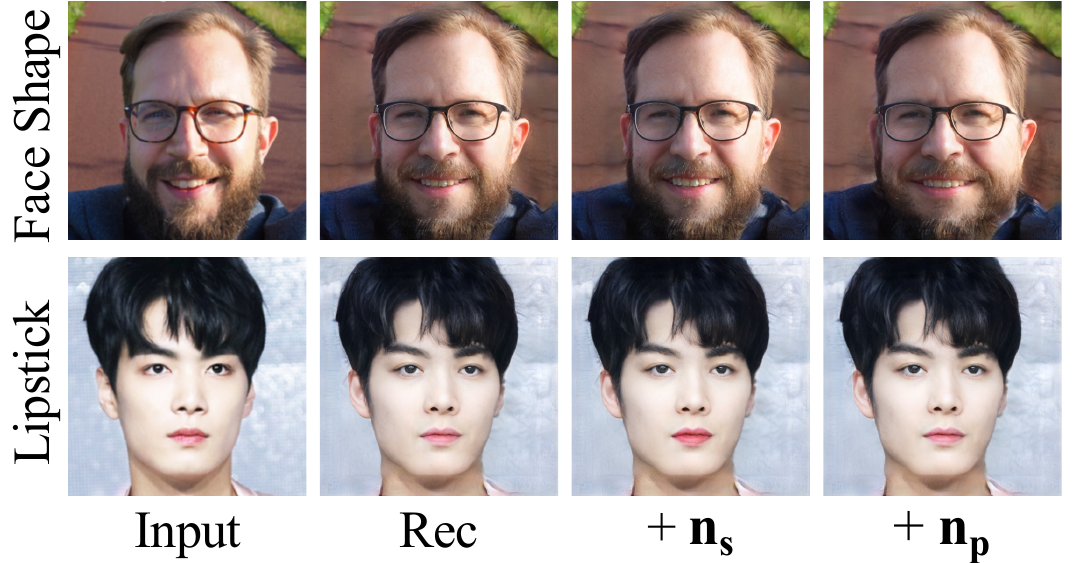}
        \caption{
            Visual results on attribute editing within the latent space and the padding space
        }
        \label{fig:1shot_editing_quantify}
    \end{minipage}
    \hfill
    \begin{minipage}[b]{0.52\linewidth}
    \centering
    \scriptsize
    \setlength{\tabcolsep}{5pt}
        \begin{tabular}{c|ccc}
            \toprule
            \textbf{Attribute} &   $\mathbf{MSE_{p}}$ & $\mathbf{MSE_{s}}$ &  
            \textbf{Factor} \\
            \midrule
            face shape     &  116$\times$1e-4  &  9$\times$1e-4  &  12.89:1  \\
            bangs          &  471$\times$1e-4  &  65$\times$1e-4 &  7.25:1   \\
            glasses        &  245$\times$1e-4  &  42$\times$1e-4 &  5.83:1   \\
            \midrule
            lipstick       &  0.437$\times$1e-4  &  6$\times$1e-4  & 1:13.73 \\
            eyebrow        &  0.880$\times$1e-4  &  14$\times$1e-4 & 1:15.90 \\
            \bottomrule
            \multicolumn{4}{c}{} \\
        \end{tabular}
        \captionof{table}{
            Qualitative comparisons on latent editing and padding editing.
            ``Factor'' indicates the ratio of
            $\mathit{MSE_{p}}$ to $\mathit{MSE_{s}}$
        }
        \label{tab:1shot_editing_quantify}

    \end{minipage}
\end{figure}

\begin{figure}[t]
    \centering
    \includegraphics[width=0.79\linewidth]{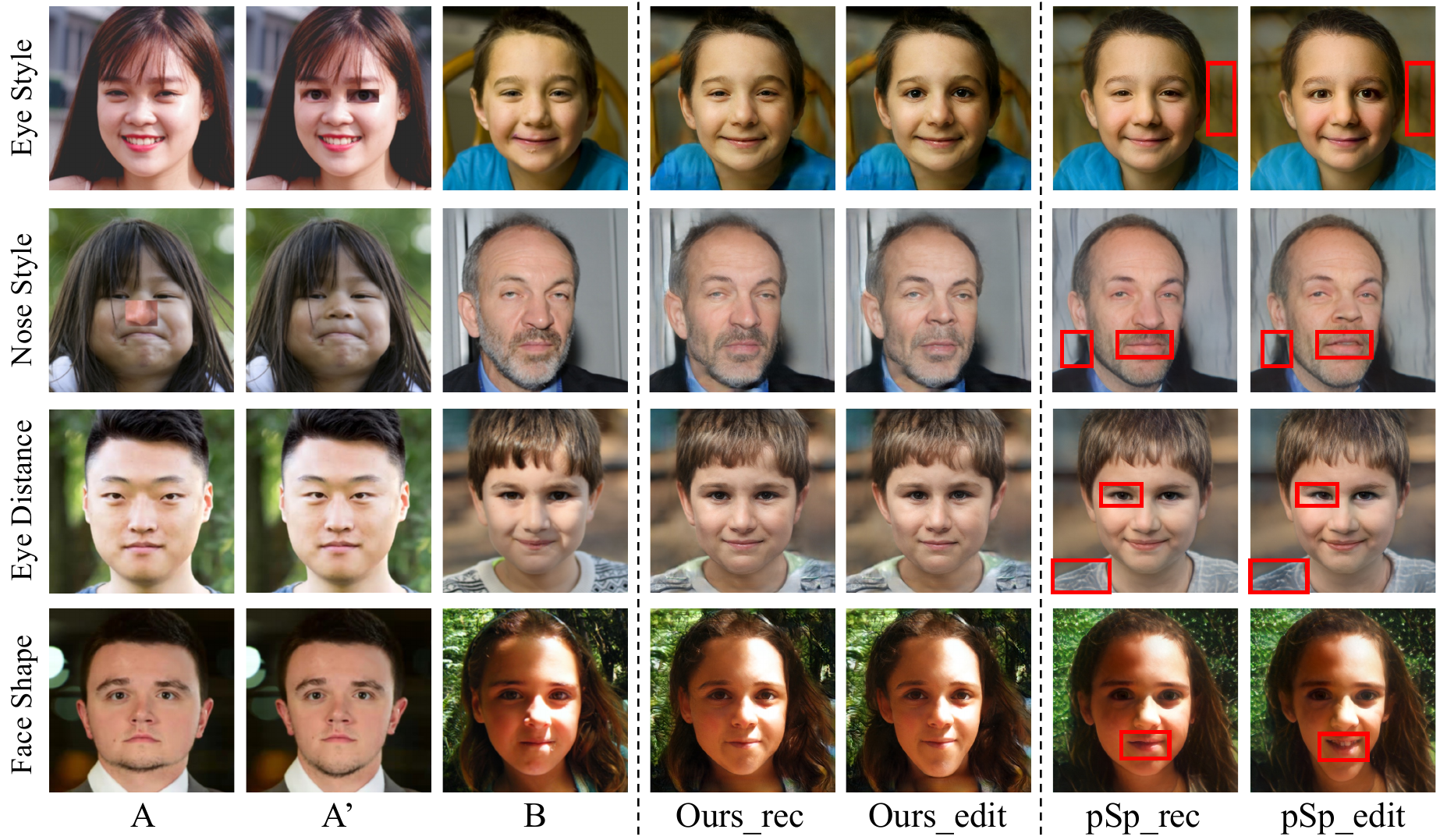}
    \caption{
        \textbf{Customized manipulation} defined by \textit{only one} image pair (A and A').
        Compared to pSp~\cite{richardson2021pSp}, which introduces unwanted changes to the target image (B), our approach presents a more precise editing performance
    }
    \label{fig:1shot_editing_comparision}
\end{figure}

\noindent\textbf{Evaluation and Analysis.}
We firstly explore the effects of $\mathbf{n_{p}}$ and $\mathbf{n_{s}}$ with respect to various attributes. 
As shown in \cref{fig:1shot_editing_quantify}, for a spatial attribute such as face shape, editing with $\mathbf{n_{p}}$ works fine while $\mathbf{n_{s}}$ rarely affects the image.
However, $\mathbf{n_{s}}$ works well for the style-related attributes like lipstick while the editing effect brought by $\mathbf{n_{p}}$ can be ignored.
To quantitatively evaluate the editing effects of $\mathbf{n_{p}}$ and $\mathbf{n_{s}}$, $MSE_{p}$ and $MSE_{s}$ between the inversion and editing results in the non-edited region like~\cite{zhu2021lowrankgan} are calculated and reported.
Specifically, $MSE_{p}$ is calculated over the 50 randomly-sampled faces and their editing results driven by $\mathbf{n_{p}}$.
$MSE_{s}$ is obtained in a similar manner with $\mathbf{n_{s}}$.
\cref{tab:1shot_editing_quantify} presents the quantitative results of semantic directions in different space.  
The factor between $MSE_{p}$ and $MSE_{s}$ is also reported for better comparison.
We discover that $\mathbf{n_{p}}$ influences target faces far more than $\mathbf{n_{s}}$ for spatial attributes such as face shape and bangs.
When it comes to style-related attributes like lipstick and brows, $\mathbf{n_{s}}$ appears to change the inversion images much more than $\mathbf{n_{p}}$.
Given this property, we can only use the $\mathbf{n_p}$ or the $\mathbf{n_s}$ for spatial or style-related editing, respectively.
The use of semantic directions separately is extremely beneficial in avoiding unwanted changes to the inversion result.
\cref{fig:1shot_editing_comparision} presents the editing results of $\mathbf{n_p}$ and $\mathbf{n_s}$.
We also include the one-shot editing results of pSp, which can be achieved in $\W+$ space.
Naively computing latent difference and applying these directions in $\W+$ space with pSp often interferes with other unconcerned attributes.
For example, a change to eye or nose style can be accompanied by a change in the background.
Besides, the face shape direction in $\W+$ space produces a larger smile in the portrait, whereas ours does not affect any attributes other than face shape.
Additional customized manipulation results of other attributes are provided in \supp.

\section{Conclusion}\label{sec:conclusion}
In this work, we attribute the unsatisfactory GAN inversion performance to inductive bias brought by zero padding. 
Thus we propose padding space ($\P$ space) to complement $\W+$ space for the reconstruction of spatial details, particularly for the out-of-distribution objects.
A carefully designed encoder is further proposed to learn the latent code and the padding coefficients jointly.
The experiments demonstrates that the padding coefficients can encode rich spatial information such as pose and contour and thus our method enables two novel applications, \textit{i.e.}, face blending and one-shot customized editing.
However, the distribution gap between the learned paddings and original features occasionally causes artifacts at the edge.
This phenomenon can be largely relieved by adversarial training.
We strongly oppose the abuse of our method in violating privacy and security. 
We hope it can be used to improve the fake detection systems.

\noindent\textbf{Acknowledgement.} This work was supported in part by the National Natural Science Foundation of China (Grant No. 61991450) and the Shenzhen Key Laboratory of Marine IntelliSense and Computation (ZDSYS20200811142605016). We thank Zhiyi Zhang for the technical support.

\bibliographystyle{splncs04}
\bibliography{ref}
\clearpage

\appendix
\section*{Appendix}
\section{Overview}\label{appendix:sec:overview}

This supplementary material is organized as follows.
\cref{appendix:sec:implement} describes the detailed architecture of our encoder.
\cref{appendix:sec:ablation} verifies how the regularization loss, $\mathcal{L}_{reg}$, helps improve the editing performance.
\cref{appendix:sec:property} further validates the property of the proposed padding space with interpolation results, as the supplement to Sec.~4.3 of the submission.
\cref{appendix:sec:results} provides more visual results on inversion, face blending, and customized manipulation.

\section{Encoder Structure}\label{appendix:sec:implement}

\cref{appendix:tab:arch} provides the detailed architecture of our encoder, by taking a 18-layer StyleGAN~\cite{stylegan, stylegan2} generator as an instance. 
Note that input images are resized to $256\times256$ at first.
Features are first extracted from the backbone and refined by FPN. %
Then style latent codes of $\W+$ space are obtained with Map2Style borrowed from~\cite{richardson2021pSp}. 
Paddings are obtained by convolving the $512\times32^{2}$ FPN feature map with the Resblocks and $1\times1$ convolutions (denoted as ``Padding Extracting Convolutions'').
The ``Style Codes'' term indicates which layer of StyleGAN the style latent codes will be sent to.
Additionally, we equip the Resblocks with squeeze and excitation block~\cite{roy2018seblock} to enhance them.

\begin{table*}[!ht]
  \caption{
    \textbf{Encoder structure} based on ResNet-IR-50~\cite{resnet, he2016resnetir}.
    The numbers in brackets indicate the dimension of features at each level.
    Paddings are obtained by convolving the $512\times32^{2}$ FPN feature map with the Resblock and $1\times1$ convolutions (denoted as ``Padding Extracting Convolutions'')
  }
  \label{appendix:tab:arch}
  \centering\tiny
  \setlength\tabcolsep{2pt}
  \begin{tabular}{ccccccc}
    \toprule
    Stage & Backbone &   Output Shape & FPN  & Map2Style & Style Codes & Padding Extracting Convolutions  \\
    \midrule
    \multirow{2}{*}{input} &
    \multirow{2}{*}{$-$} &
    \multirow{2}{*}{$3\times 256^2$} &
    & & &  \\
    & & & & & &  \\
    \midrule
    \multirow{2}{*}{conv$_1$} &
    \multicolumn{1}{c}{3$\times$3, {64}} &
    \multirow{2}{*}{$64\times 256^2$} &
    & & &  \\
    & stride 1, 1 & & & & &  \\
    \midrule
    \multirow{3}{*}{res$_2$} &
    \blocks{64}{{64}}{3} &
    \multirow{3}{*}{$64\times 128^2$} &
    & & &  \\
    & & & & & &  \\
    & & & & & &  \\
    \midrule
    \multirow{3}{*}{res$_3$} &
    \blocks{{128}}{{128}}{4} &
    \multirow{3}{*}{128$\times$ $64^2$} &
    \multirow{3}{*}{512$\times$ $64^2$} &
    \multirow{3}{*}{11$\times$512} &
    \multirow{3}{*}{\shortstack{
    \\
    Layer 8-18  \\[-2.2pt]
    }}  
    &
    \\
    & & & & & &  \\
    & & & & & &  \\
    \midrule
    \multirow{3}{*}{res$_4$} &
    \blocks{{256}}{{256}}{14} &
    \multirow{3}{*}{256$\times$ $32^2$} &
    \multirow{3}{*}{$512\times 32^2$} &
    \multirow{3}{*}{4$\times$512} &
    \multirow{3}{*}{\shortstack{
        \\
        Layer 4-7  \\[-2.2pt]
    }}
    &
    \multirow{3}{*}{
    \(\left[\begin{array}{c}\text{1$\times$1, 512}\\[-.1em] \text{3$\times$3, 512}\\[-.1em] \text{3$\times$3, 512}\end{array}\right]\)$\times$2
    }
    \multirow{3}{*}{\&}
    \multirow{3}{*}{
    \(\left[\begin{array}{c}
    \text{1$\times$1, 512}
    \end{array}\right]\)$\times$4
    }
    \\
    & & & & & &   \\
    & & & & & &   \\
    \midrule
    \multirow{3}{*}{res$_5$} &
    \blocks{{512}}{{512}}{3} &
    \multirow{3}{*}{$512\times 16^2$} &
    \multirow{3}{*}{$512\times 16^2$} &
    \multirow{3}{*}{3$\times$512} & 
    \multirow{3}{*}{\shortstack{
    \\
    Layer 1-3 \\[-0.2pt]
    }} 
    &
    \\
    & & & & & &  \\
    & & & & & &  \\
    \bottomrule
  \end{tabular}
\end{table*}

\section{Ablation Study on Regularization Loss}\label{appendix:sec:ablation}

\begin{figure}[t]
    \centering
    \includegraphics[width=0.80\linewidth]{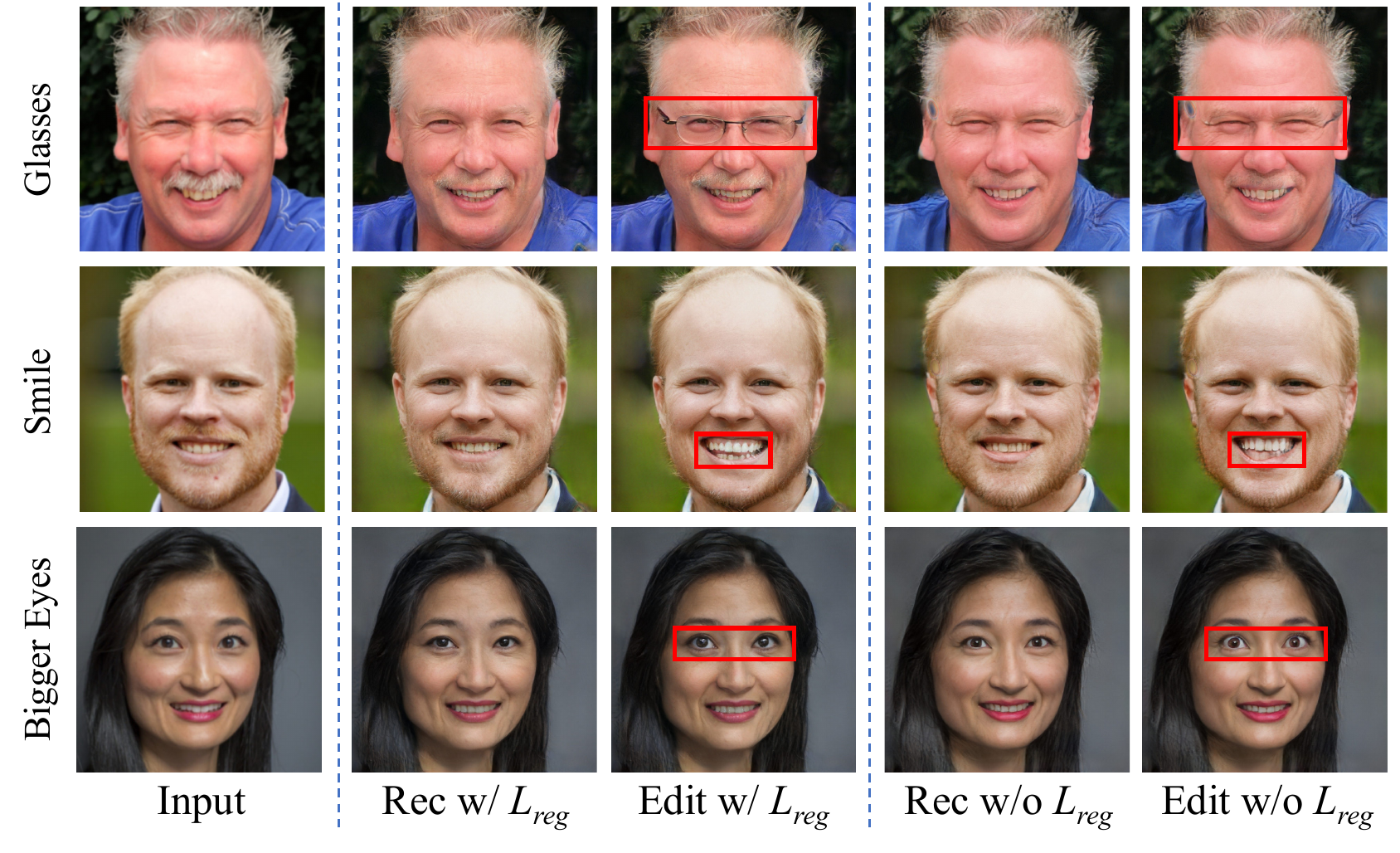}
    \caption{
        Ablation study on $L_{reg}$. Models trained with $L_{reg}$ have better performance in editing while show slightly inferior performance in reconstruction
    }
    \label{appendix:fig:ablation_lreg}
\end{figure}

Recall that we introduce the regularization loss $\mathcal{L}_{reg}$ in Eq.(5) of the submission to enhance the encoder in editing ability.
In order to achieve better editing performance, this regularization is applied to encourage the inverted code in $\W+$ space to be subject to the native latent distribution.

In this section we ablate $\mathcal{L}_{reg}$ on editing tasks.
We separately learn two proposed encoders with or without $\mathcal{L}_{reg}$ and use them to invert the input faces.
Then off-the-shelf semantic directions from~\cite{shen2020interfacegan, zhu2021lowrankgan} are adopted to edit the inversion results.
As demonstrated in~\cref{appendix:fig:ablation_lreg}, during editing the model trained without $\mathcal{L}_{reg}$ produces inferior results such as failed editing for eye-glasses, blurred teeth for smile, and unnatural expression for bigger eyes.
While models trained with $L_{reg}$ have better performance in editing but show slightly inferior performance in reconstruction.

\section{Property of Padding Space}\label{appendix:sec:property}

\begin{figure}[t]
    \centering
    \includegraphics[width=1.0\linewidth]{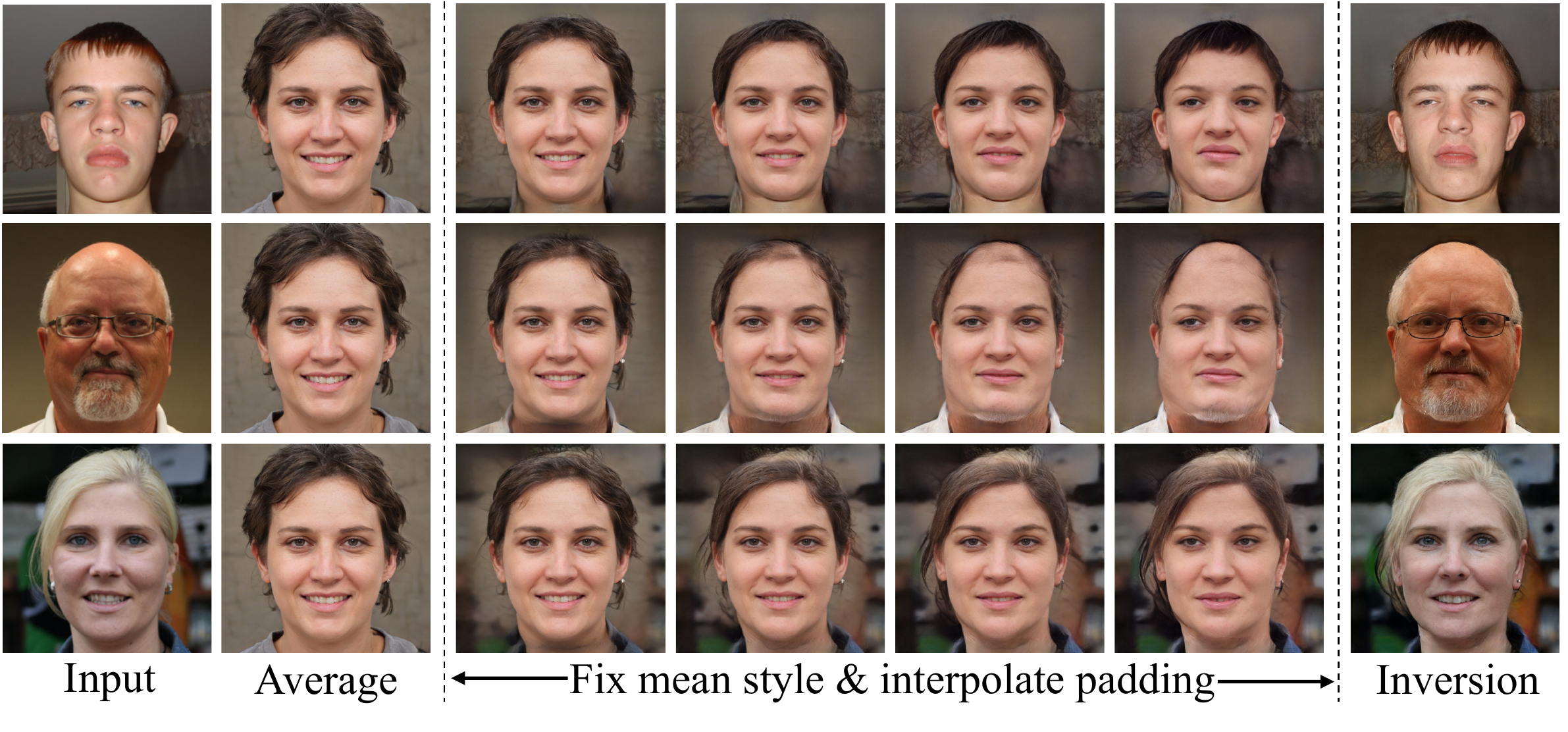}
    \caption{
        \textbf{Analysis of the padding effect} on FFHQ~\cite{stylegan} test set.
        We fix the latent code as the statistical average and interpolate the padding from the fixed constants in the generator to the coefficients specifically learned for inversion.
        It verifies that padding encodes the spatial information
    }
    \label{appendix:fig:face_inter1Pos}
\end{figure}

\begin{figure}[!ht]
    \centering
    \includegraphics[width=1.0\linewidth]{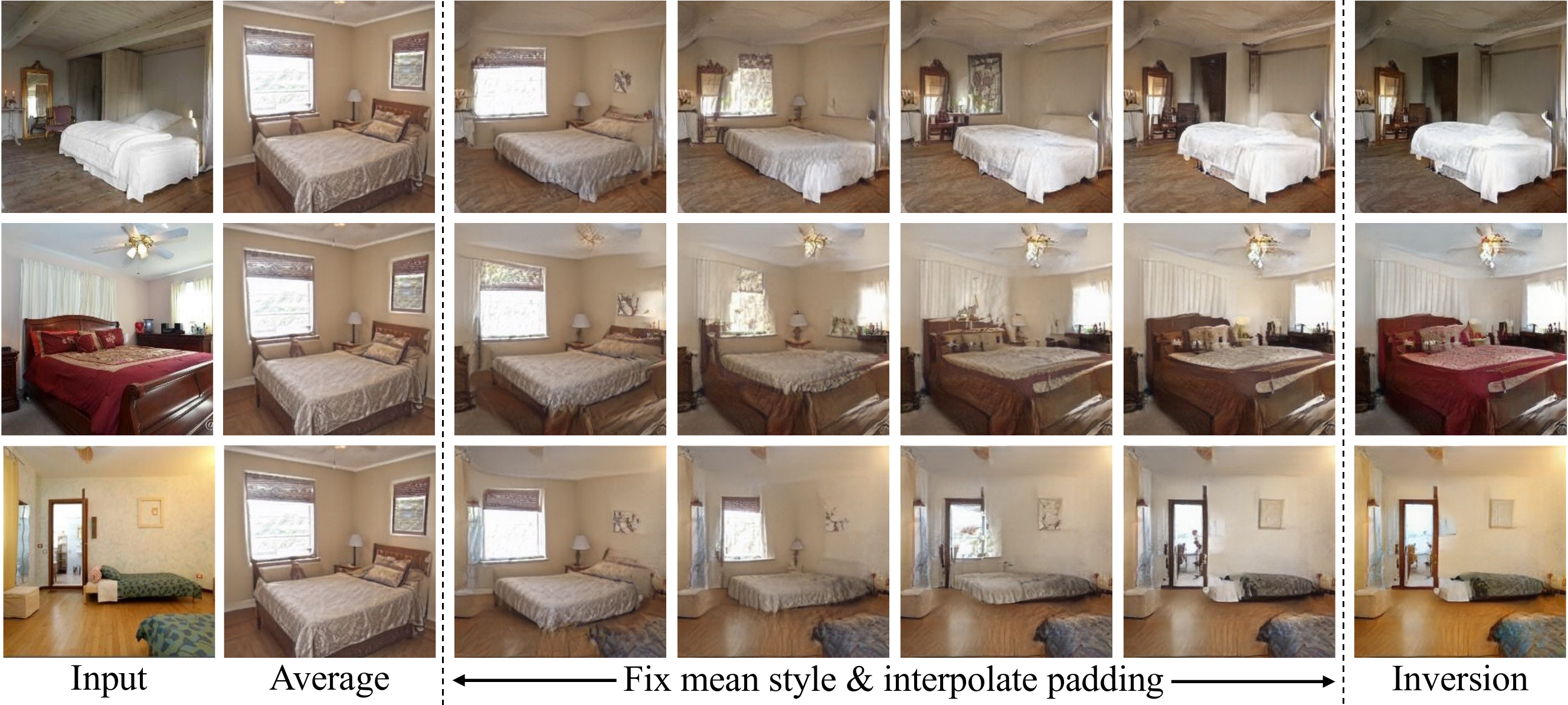}
    \caption{
        \textbf{Analysis of the padding effect} on LSUN Bedroom~\cite{yu2015lsun}.
        We fix the latent code as the statistical average and interpolate the padding from the fixed constants in the generator to the coefficients specifically learned for inversion
    }
    \label{appendix:fig:bedroom_inter1Pos}
\end{figure}

\begin{figure}[t]
    \centering
    \includegraphics[width=1.0\linewidth]{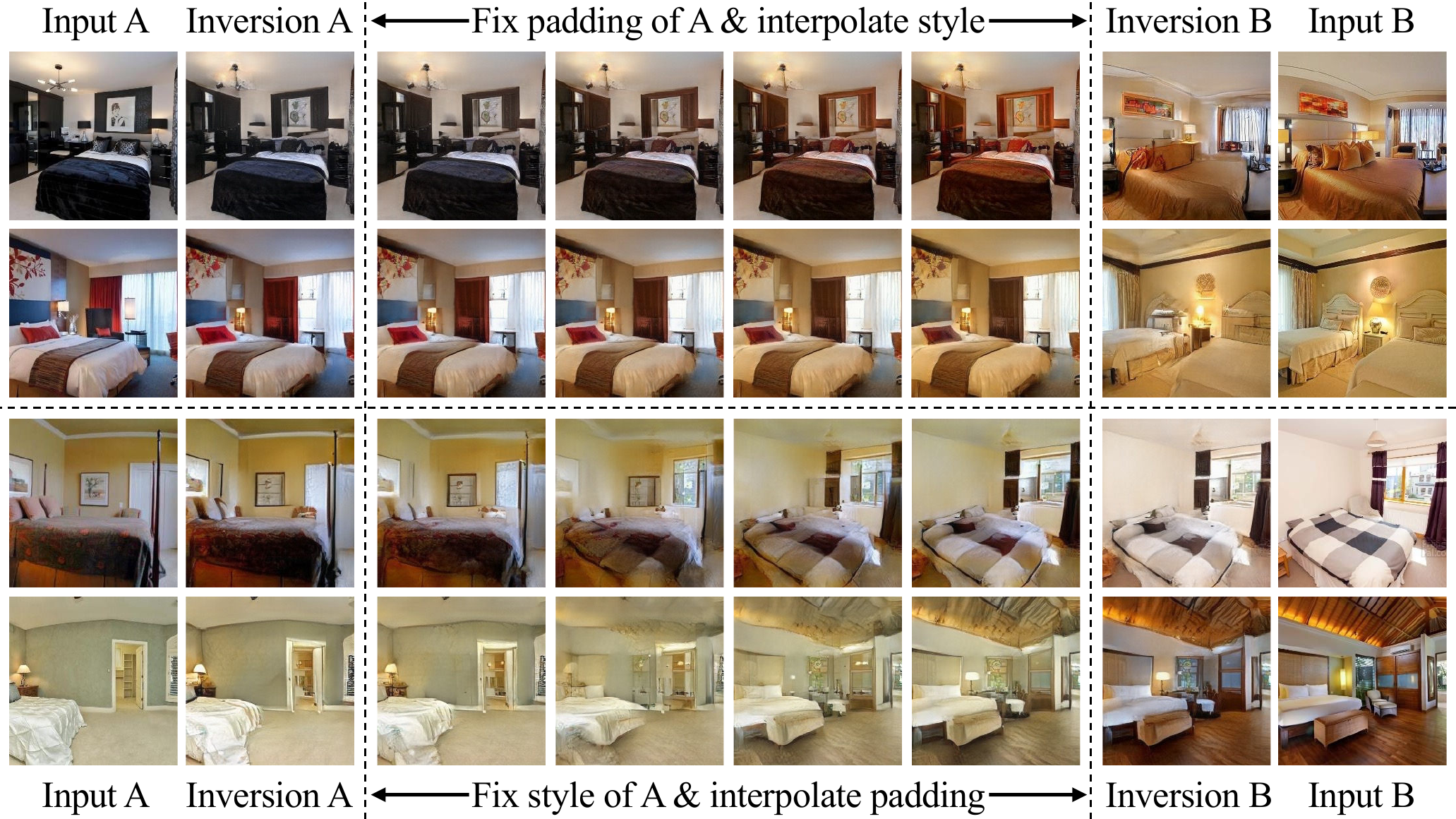}
    \caption{
        \textbf{Analysis of the extended inversion space} on LSUN Bedroom~\cite{yu2015lsun}.
        We perform interpolation both in the latent space and in the padding space
    }
    \label{appendix:fig:bedroom_inter2}
\end{figure}

Recall that we conduct interpolation experiments in Sec~4.3 of the submission. 
Here we additionally provide the interpolation results on the FFHQ test set (the spared last 5k images) and LSUN Bedroom test set to validate the property of the padding space ($\P$ space).

Firstly, we fixed the style latent codes as the average and interpolate the paddings between the fixed constants and coefficients extracted by the encoder.
From \cref{appendix:fig:face_inter1Pos} and \cref{appendix:fig:bedroom_inter1Pos} we can conclude that during interpolation the style information such as face details and color style is maintained.
While the pose and contour keep changing from the average to the input image.
Then as in Sec~4.3 of the submission, we perform interpolation between two inversion results.
Specifically, we first invert two real images A and B to $\P$ and $\W+$ space.
Then we fix one of the paddings or latent codes and interpolate the other.
As in \cref{appendix:fig:bedroom_inter2}, when padding coefficients are kept to be fixed, the style varies smoothly when the latent code changes.
Also when style latent codes are fixed, the spatial information changes from the image A to B while the image style stays fixed.
In summary, these results indicate that the paddings encode spatial information such as the contour of the face, background, and pose. 

\section{Additional Results}\label{appendix:sec:results}

In this section, we provide more visualization results of inversion, face blending and customized editing.

\subsection{Inversion}\label{appendix:subsec:inversion}

\begin{figure}[!ht]
    \centering
    \includegraphics[width=1.0\linewidth]{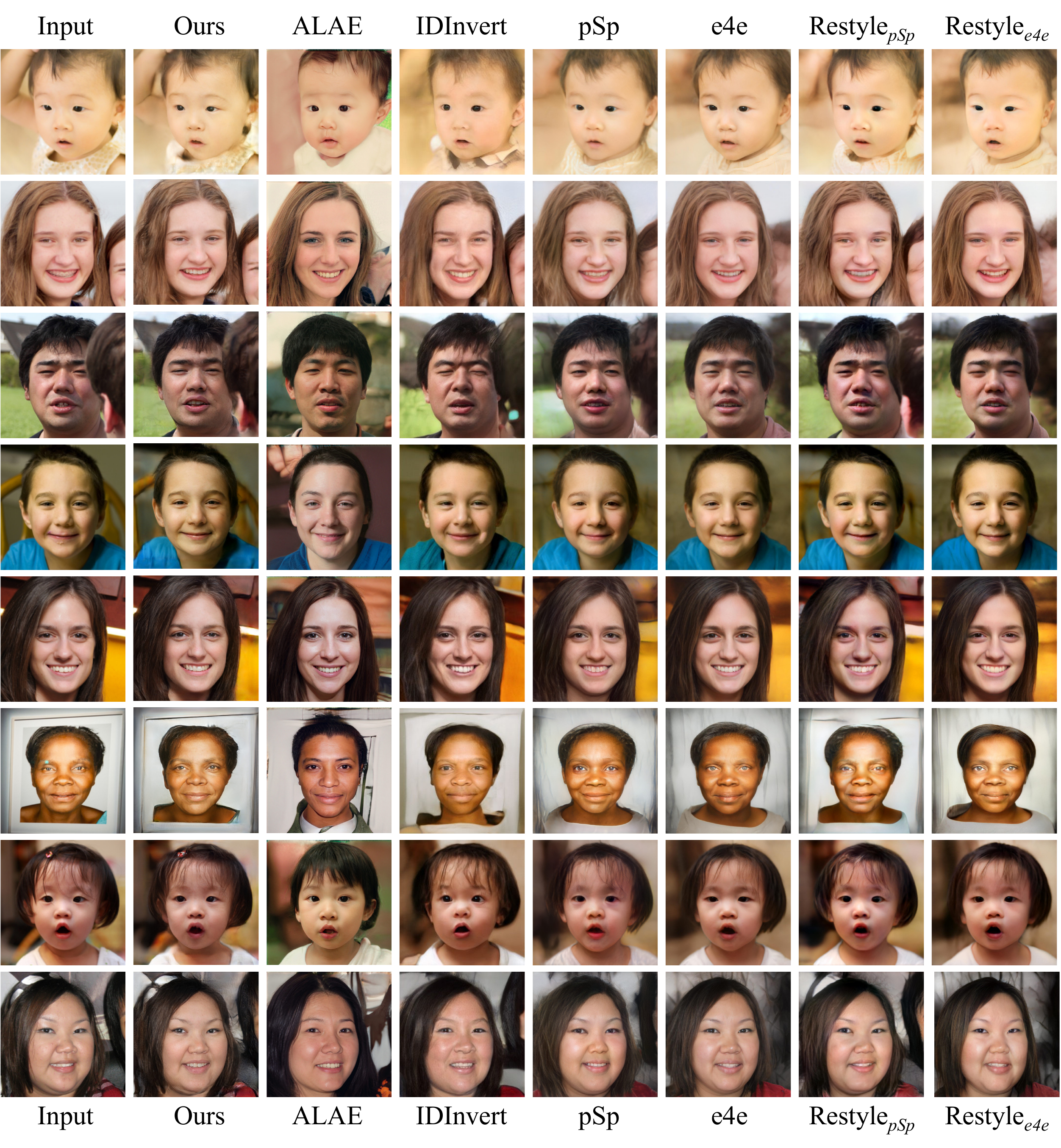}
    \caption{
        Additional inversion results of various methods on FFHQ~\cite{stylegan} test set.
        Note that our method perform better when reconstructing out-of-distribution spatial details such as the arm, secondary face, background, hairpins, and head-wears
    }
    \label{appendix:fig:inversion_more_ffhq}
\end{figure}

\begin{figure}[!ht]
    \centering
    \includegraphics[width=1.0\linewidth]{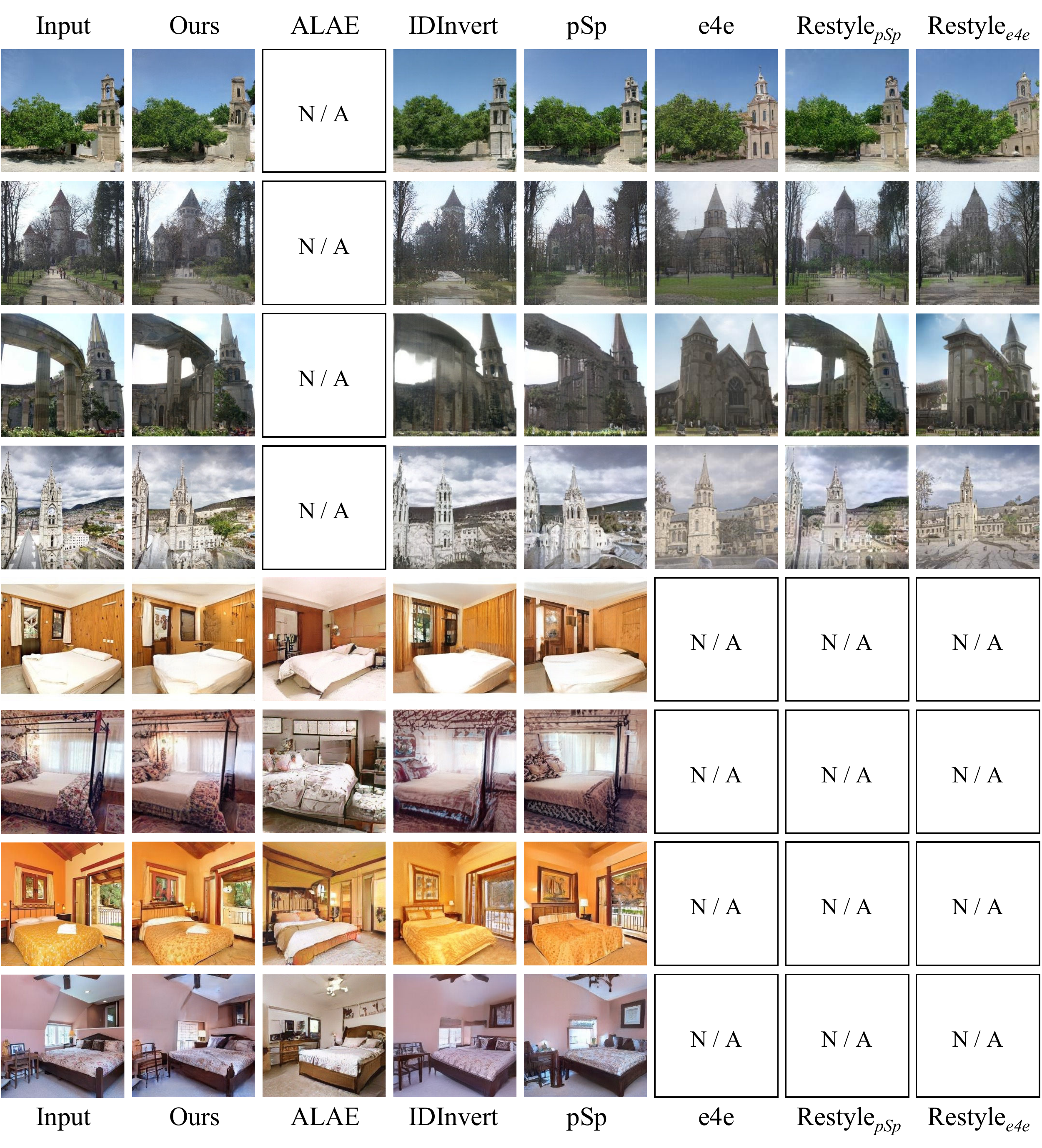}
    \caption{
         Additional inversion results of various methods on LSUN~\cite{yu2015lsun} Church and Bedroom test set. 
         Our method perform better when reconstructing spatial details such as and holes and pillows. N/A indicates the pretrained model is not available
    }
    \label{appendix:fig:inversion_more_LSUN}
\end{figure}

\begin{figure}[!ht]
    \centering
    \includegraphics[width=1.0\linewidth]{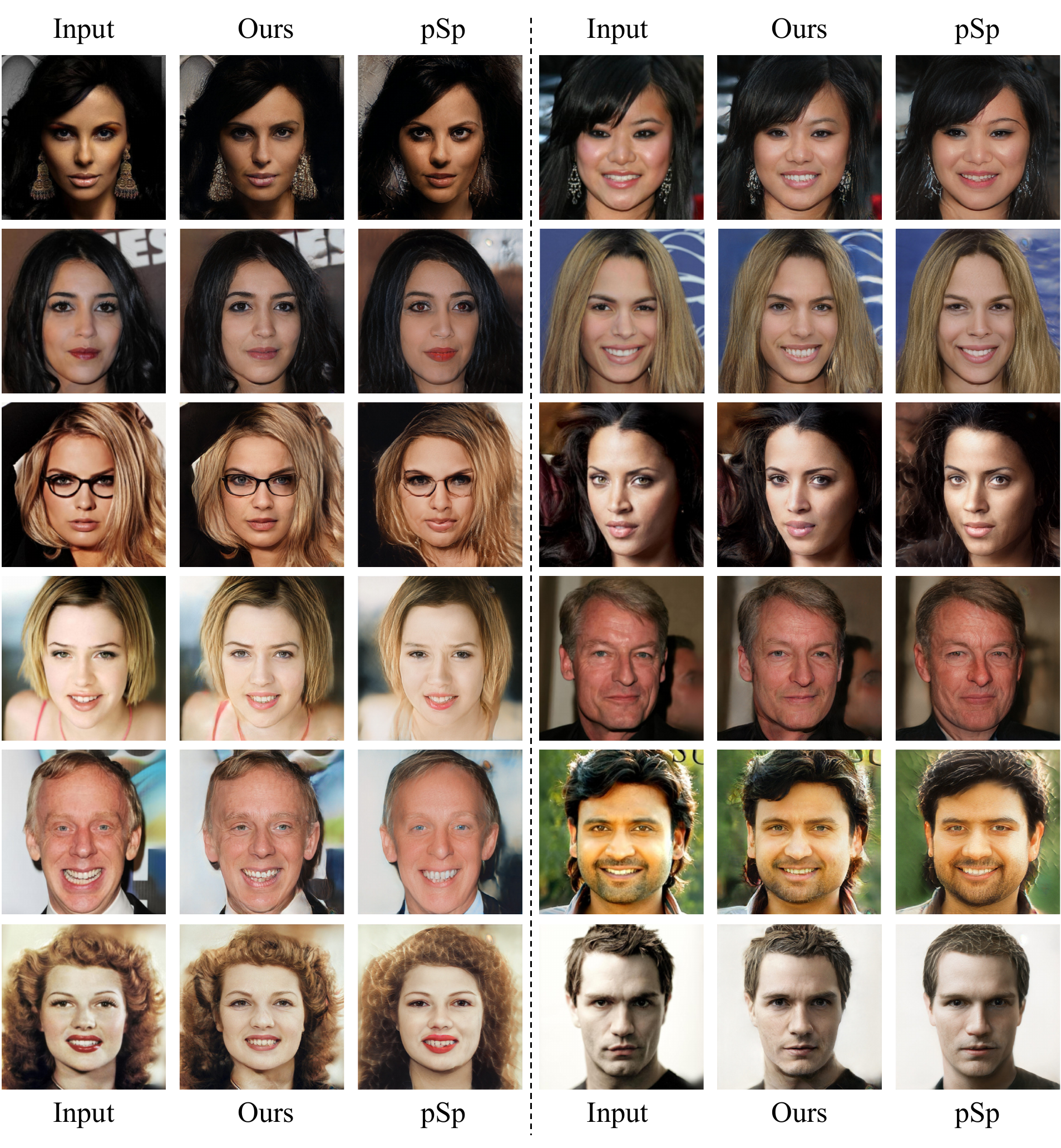}
    \caption{
        Additional inversion comparisons between our method and the baseline pSp~\cite{richardson2021pSp} on CelebA-HQ~\cite{progan, liu2015deep} test set with pretrained StyleGANv1. 
        Our method has better performance in reconstruction of spatial details such as earring, background, and haircut
    }
    \label{appendix:fig:inversion_more_celeba}
\end{figure}

Here we qualitatively compare our inversion results with the ones of ALAE~\cite{pidhorskyi2020alae}, IDInvert~\cite{zhu2020idinvert}, pSp~\cite{richardson2021pSp}, e4e~\cite{tov2021e4e}, and Restyle~\cite{alaluf2021restyle}.
\cref{appendix:fig:inversion_more_ffhq} and \cref{appendix:fig:inversion_more_LSUN} respectively demonstrate inversion results on FFHQ~\cite{stylegan} and LSUN~\cite{yu2015lsun} test set, from which we conclude that our method performs better in reconstruction of spatial details. 
We additionally validate the effectiveness of the proposed padding space on StyleGAN1~\cite{stylegan}.
\cref{appendix:fig:inversion_more_celeba} demonstrate the inversion results of our method and the baseline pSp with pretrained StyleGAN1 on CelebA-HQ~\cite{progan, liu2015deep} test set.

\subsection{Face Blending }\label{appendix:subsec:face_blending}

\begin{figure}[!ht]
    \centering
    \includegraphics[width=1.0\linewidth]{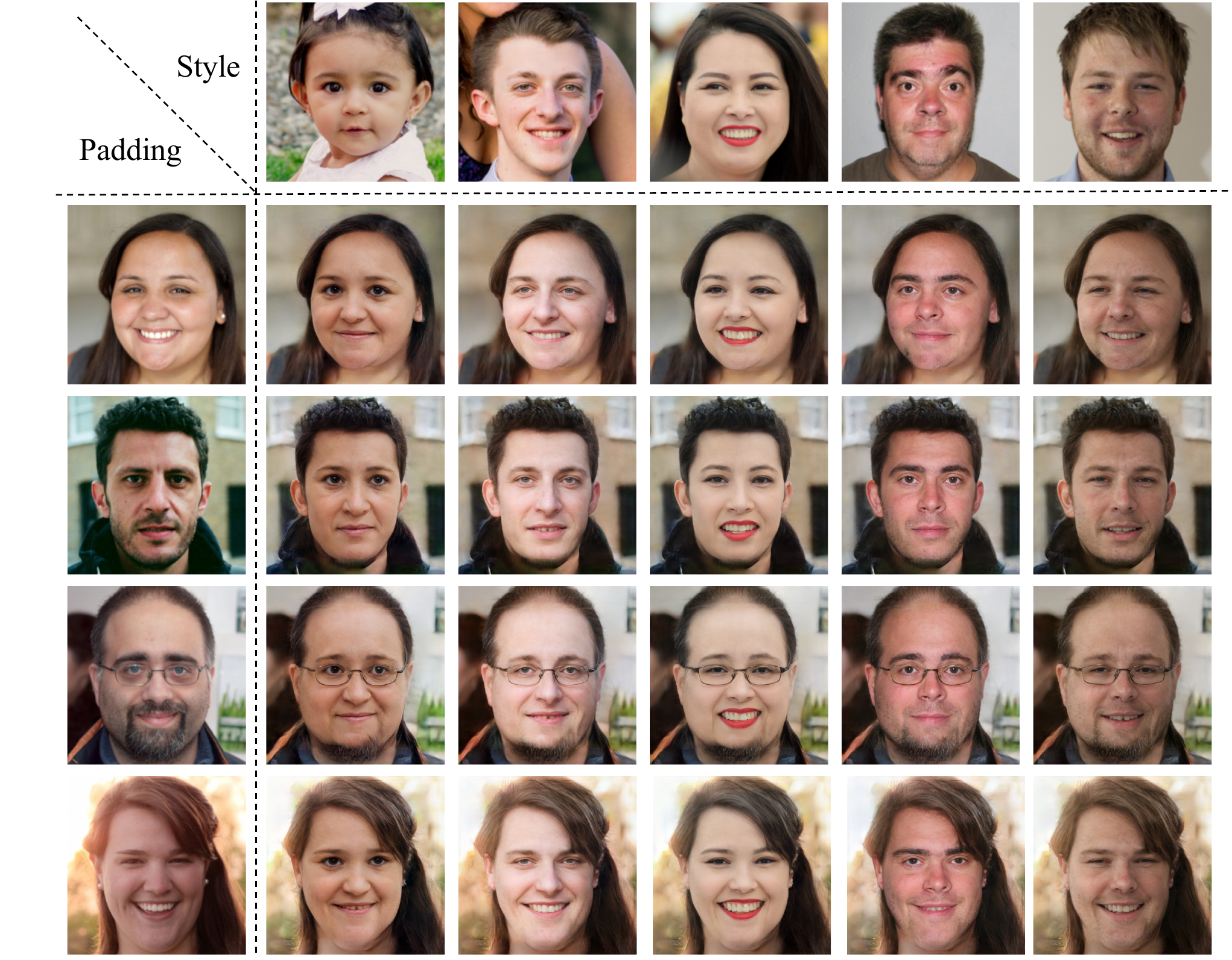}
    \caption{
        Additional face blending results by combining different paddings and style latent codes, as described in Sec~4.3 of the submission. 
        We can conclude that the paddings encode spatial information such as face pose and contour while style latent codes encode face details
    }
    \label{appendix:fig:face_blending_more}
\end{figure}

Recall that our method enables face blending by simply swapping the style latent codes of two input faces.
More face blending results are provided in in~\cref{appendix:fig:face_blending_more}.
We can conclude that the paddings encode spatial information such as the face pose and contour while style latent codes encode face details.

\subsection{Customized Editing}
\label{appendix:subsec:customized_editing}

\begin{figure}[!ht]
    \centering
    \includegraphics[width=1.0\linewidth]{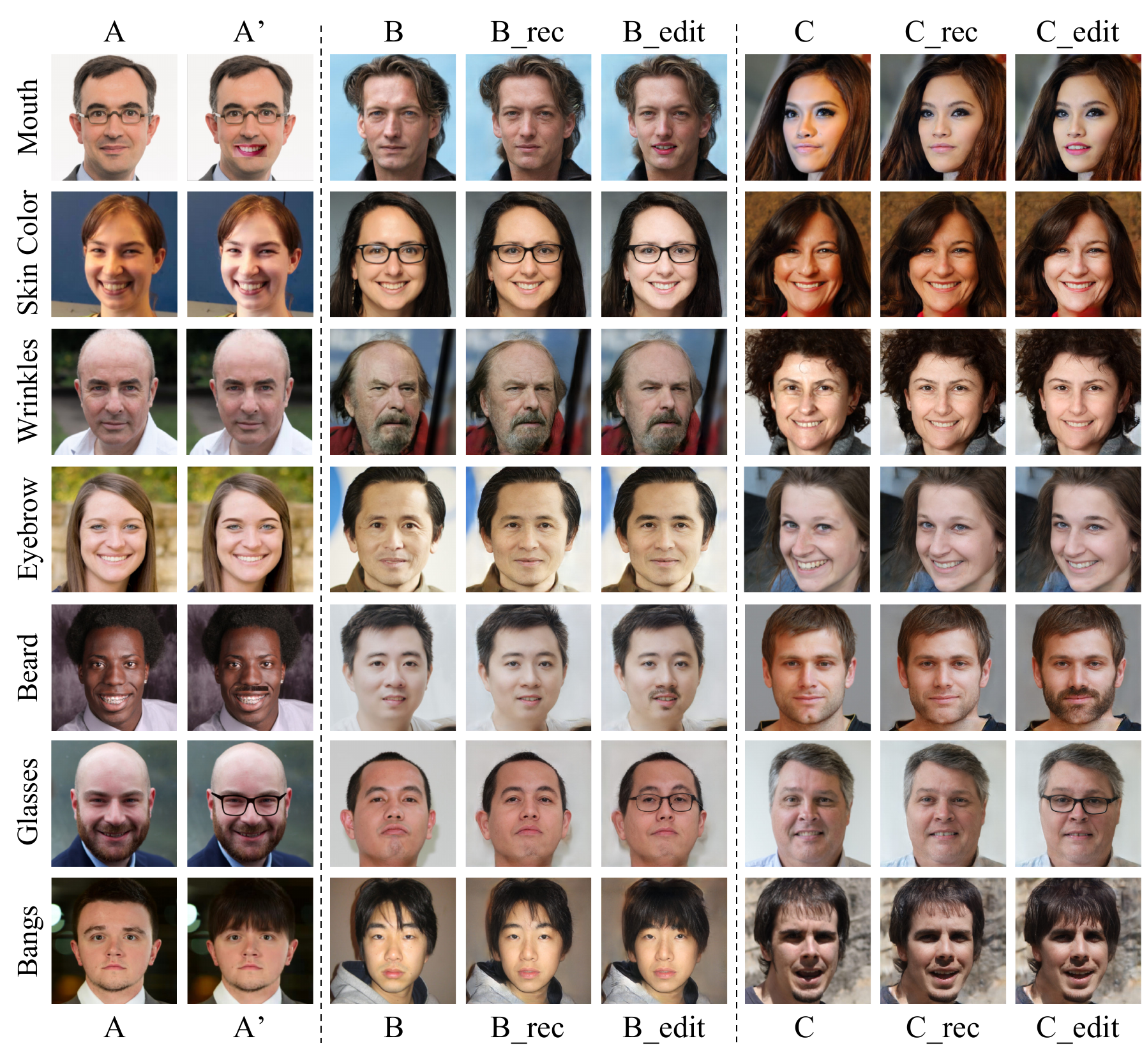}
    \caption{
        Additional results of customized editing with one pair of images.
        The semantic direction defined by one pair of images (A and A') can be applied to arbitrary samples (like B and C) to precisely edit them. 
        ``rec'' denotes the inversion result of the input.
        Among the attributes, glasses and bangs are spatial-related thus we edit them only in the padding space and others are style-related and are edited in $\W+$ space, as described in Sec~4.4 of the submission
    }
    \label{appendix:fig:1ShotEdit_more}
\end{figure}

Recall that we define semantic directions with paired images and edit the inversion results with them in Sec~4.4 of the submission.
Here we provide more customized editing results in~\cref{appendix:fig:1ShotEdit_more}.
We can conclude that the semantic direction defined by one pair of images (A and A') can be applied to arbitrary samples (such as B and C) to precisely edit them.

\end{document}